\documentclass{article}

\usepackage{arxiv}

\usepackage[utf8]{inputenc} 
\usepackage[T1]{fontenc}    
\usepackage{hyperref}       
\usepackage{url}            
\usepackage{booktabs}       
\usepackage{amsfonts}
\usepackage{amsmath}
\usepackage{nicefrac}       
\usepackage{microtype}      
\usepackage{lipsum}
\usepackage{graphicx}
\usepackage{caption}
\usepackage{subcaption}
\usepackage[numbers]{natbib}
\usepackage{tikz}
\usetikzlibrary{3d,arrows.meta,positioning}
\usetikzlibrary{positioning}
\graphicspath{{Fig/}}
\usepackage{algorithm}
\usepackage{algcompatible}

\title{Fraud is Not Just Rarity: A Causal Prototype Attention Approach to Realistic Synthetic Oversampling}

\author{
 \href{https://orcid.org/0009-0001-1284-6631}{\includegraphics[scale=0.06]{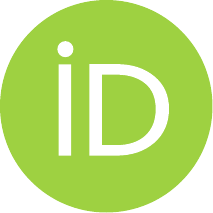}\hspace{1mm}Claudio Giusti} \\
  Department of Mathematics and Computer Science\\
  University of Catania\\
  Catania, CT 95125 \\
  \texttt{claudio.giusti@studium.unict.it} \\
   \And
  \href{https://orcid.org/0000-0001-8315-351X}{\includegraphics[scale=0.06]{orcid.pdf}\hspace{1mm}Luca Guarnera} \\
  Department of Mathematics and Computer Science\\
  University of Catania\\
  Catania, CT 95125 \\
  \texttt{luca.guarnera@unict.it} \\
  \And
  \href{https://orcid.org/0000-0001-6975-2241}{\includegraphics[scale=0.06]{orcid.pdf}\hspace{1mm}Mirko Casu} \\
  Department of Mathematics and Computer Science\\
  University of Catania\\
  Catania, CT 95125 \\
  \texttt{mirko.casu@phd.unict.it} \\
  \And
  \href{https://orcid.org/0000-0001-6127-2470}{\includegraphics[scale=0.06]{orcid.pdf}\hspace{1mm}Sebastiano Battiato} \\
  Department of Mathematics and Computer Science\\
  University of Catania\\
  Catania, CT 95125 \\
  \texttt{sebastiano.battiato@unict.it} \\
}
\renewcommand{\shorttitle}{Fraud is Not Just Rarity}

\begin{document}
\maketitle
\begin{abstract}
Detecting fraudulent credit card transactions remains a significant challenge, due to the extreme class imbalance in real-world data and the often subtle patterns that separate fraud from legitimate activity. Existing research commonly attempts to address this by generating synthetic samples for the minority class using approaches such as GANs, VAEs (Variational Autoencoders), or hybrid generative models. However, these techniques, particularly when applied only to minority-class data, tend to result in overconfident classifiers and poor latent cluster separation, ultimately limiting real-world detection performance.
In this study, we propose the Causal Prototype Attention Classifier (CPAC), an interpretable architecture that promotes class-aware clustering and improved latent space structure through prototype-based attention mechanisms and we couple it with the encoder of a Variational Autoencoder–Generative Adversarial Network (VAE-GAN) in order to achieve improved latent cluster separation moving beyond post-hoc sample augmentation.
We compared CPAC-augmented models to traditional oversamplers, such as SMOTE, as well as to state-of-the-art generative models, both with and without CPAC-based latent classifiers. Our results show that classifier-guided latent shaping with CPAC delivers superior performance, achieving an F1-score of 93.74\% and recall of 92.85\%, along with improved latent cluster separation. Further ablation studies and visualizations provide deeper insight into the benefits and limitations of classifier-driven representation learning for fraud detection. The codebase for this work can be found at the following link: \url{https://github.com/claudiunderthehood/VAEGAN-CPAC.git}.
\end{abstract}

\section{Introduction}
The escalation of cyber threats has made anomaly detection central in computer security. Organizations face increasingly sophisticated attacks, from targeted intrusions and APTs to advanced fraud schemes~\cite{sommer2010outside, garcia2014empirical, bhuyan2014network}. Malicious actions are rare and often hidden within massive volumes of legitimate activity, making minority-class detection one of the key challenges. Automated systems must identify new threats without excessive false positives, while ensuring interpretability for compliance~\cite{axelsson2000base, chandola2009anomaly, bridges2019machine}.
Large-scale analyses~\cite{romanosky2016examining} highlight that most cyber incidents are not headline-grabbing, but sector risks differ greatly. Fraud types keep evolving, including affiliate marketing abuse~\cite{snyder2016characterizing}, and user response is shaped by liability and reimbursement differences across countries~\cite{becker2018international}.
Detecting frauds in financial settings~\citep{west2016, edge2009}, especially in systems based on rule induction~\cite{vorobyev2022} where it is rare, remains a central challenge.  Similar issues arise in domains like deepfake detection~\cite{ke2025detection,amerini2025deepfake,sharma2024banking}. New biases such as ‘impostor bias’~\cite{CASU2024301795} further complicate anomaly detection.
Traditional machine learning models such as Logistic Regression~\cite{hosmer2013applied}, Random Forest~\cite{breiman2001random}, and XGBoost~\citep{chen2016xgboost} are widely used, but their effectiveness drops with heavy imbalance, often requiring sophisticated sampling or cost-sensitive learning.
Among oversampling strategies, two main families dominate:
\begin{itemize}
\item \textbf{SMOTE}-based methods~\citep{chawla2002smote}: Synthesize new minority samples via interpolation, balancing the training set. They are effective but can produce redundant or overly smooth data.
\item \textbf{Generative models} (VAEs~\citep{kingma2014autoencoding}, GANs~\citep{goodfellow2014generative}, DMs~\cite{ho2020denoising,sohl2015deep}): These generate diverse samples from learned distributions, but often require significant tuning and are usually trained only on minority data, which can limit diversity and generalization.
\end{itemize}
Deep classifiers are effective for fraud detection~\citep{li2018deep}, but are often black-boxes, making interpretability difficult in sensitive domains. Generative methods typically focus on augmenting the minority class without shaping the latent space for better decision-making.
To address these limitations, we propose the Causal Prototype Attention Classifier (CPAC), a lightweight and interpretable architecture that uses prototype-based reasoning and feature attention for robust classification under imbalance. Coupling CPAC (or similar classifiers) with generative model encoders such as VAE-GAN enables latent space shaping that maximizes class separability and interpretability, outperforming SOTA oversampling methods in both clustering and detection metrics.
We validate our approach on the Kaggle Credit Card Fraud Detection~\citep{dal2015dataset} dataset, benchmarking CPAC-augmented models against traditional classifiers, SMOTE, and generative oversamplers, as well as MLP-based latent classifiers.
The main contributions of this work are as follows:
\begin{itemize}
    \item We present the CPAC, an interpretable classifier that combines prototypes and attention for reliable fraud detection under extreme class imbalance.
    \item We introduce a classifier-guided latent shaping approach by attaching a classifier to the encoder of a VAE-GAN, enforcing class-aware clustering and improving downstream classification performance.
    \item We included the CPAC to the encoder of the VAE-GAN to improve the results of the generic classifiers by exploiting the inner qualities of the CPAC.
    \item We prove and discuss how training generative models only on fraud data proves ineffective despite the high performances that these might lead the classifiers, causing overconfidence and poor representation of the actual data.
\end{itemize}
This work will be structured as such: in Section~\ref{sec:related_work} we present the current SOTA of the literature and all the works that pushed and inspired ours. In Section~\ref{sec:methodologies} we list and explain all the techniques and models that this work uses and presents. Later, in Section~\ref{sec:experiments} we explain and analyze all the results we obtained. Section~\ref{sec:ablation} explores the importance of each component in the proposed architecture by removing them and explaining why they are critical. 
In Section~\ref{sec:discussion} we analyze and motivate why current SOTA might represent a liability for fraud detection, despite the good metrics and overall performances.
Ultimately, Section~\ref{sec:conclusions} concludes the paper with some hints at some plausible future works.

\section{Related Work}
\label{sec:related_work}
Research in fraud detection and anomaly identification has evolved along several key dimensions: data-level oversampling, deep generative modeling, and the development of interpretable or explainable classifiers~\citep{explina}. Below, we review foundational and recent advances in these areas, with particular focus on techniques most relevant to the design of robust and interpretable fraud detection systems.

\subsection{Data-Level Oversampling and Generative Models}
Handling extreme class imbalance in fraud detection has long been addressed at the data‑level. Early work introduced SMOTE~\citep{Chawla2002}, with refinements and surveys over the years~\citep{Fernandez2018}. More recently, deep generative models, including VAEs, GANs, and hybrid approaches have been applied to rebalance fraud datasets. For instance, Tang et al.~\citep{Tang2025} proposed combining GANs and VAEs to simulate realistic transaction flows for anomaly detection, showing improved detection of rare fraud behaviors. Complementary surveys summarized the landscape of GAN‑based augmentation techniques for credit card fraud detection, highlighting a diversity of architectures and strategies~\citep{SurveyGAN2022}. Despite these advances, most generative methods focus on augmenting the minority class alone, which may lead to overconfident or overfit classifiers and limited cluster separation in latent space.

\subsection{Prototypes, Attention, and Explainability}
Beyond data augmentation, prototype-based networks provide intrinsic interpretability by comparing inputs to learned class exemplars. ProtoPNet~\citep{Chen2018} pioneered this for image classification, while subsequent work further quantified the visual or semantic attributes that drive similarity~\citep{Nauta2020}. ProSeNet extended prototype reasoning to sequential data~\citep{Yao2019}, but all such methods must guard against mismatches between prototypes and actual data features~\citep{Hoffmann2021}. Attention mechanisms have likewise been adopted for per-feature weighting and have been proposed as explanation tools, though their reliability as explanations remains debated~\citep{Jain2019,Wiegreffe2019}. Model-agnostic post-hoc methods such as LIME~\citep{Ribeiro2016} and SHAP~\citep{Lundberg2017} also remain popular for explaining classifier decisions.

\subsection{Recent Detection and Generative Oversampling Approaches}
Recent years have witnessed a proliferation of generative oversampling strategies and detection methods for credit card fraud detection task, with most approaches focused on synthesizing minority class (fraud) data to address the severe class imbalance typical of this domain. Below, we summarize the methodological contributions and experimental setups of recent and representative works in this area.
Rakhshaninejad et al. (2021)~\cite{rakhshaninejad2021ensemble} propose an ensemble method that uses a weighted voting system algorithm to enforce and build more reliable classifiers for detecting frauds.
Wang et al. (2022)~\cite{Wang2022} proposed the use of Unrolled Generative Adversarial Networks (Unrolled GAN) for the oversampling of fraudulent transactions. Their method, designed to overcome issues such as mode collapse in classical GANs, generates synthetic fraud samples to augment the minority class. The Unrolled GAN is explicitly trained only on the fraudulent samples, and the generated data is added to the original dataset before training downstream classifiers. Their experiments demonstrate that Unrolled GAN-based oversampling improves classification results over classical methods like SMOTE, highlighting the capacity of deep generative models to capture minority class distributions.
Ding et al. (2023)~\cite{Ding2023} present a hybrid model that combines VAE with adversarial training (VAE-GAN) to generate synthetic fraud transactions. Their model learns the distribution of the minority class (fraud) and is trained solely on fraudulent examples, which are then used to augment the training set for downstream classifiers.
Shi et al. (2025)~\cite{Shi2025_balVAE} propose a class-imbalance-aware VAE with a transformer-based attention mechanism (Bal-VAE-Attention). Their model employs a loss function with class-aware weights to better learn from minority samples, and generates synthetic frauds for augmentation after training. Unlike earlier works, their results show that employing architectural or loss-based corrections can produce more robust synthetic samples and improved downstream detection rates.
Ahmed et al. (2025)~\cite{Ahmed2025} propose a hybrid data sampling approach for credit card fraud detection by combining SMOTE with Edited Nearest Neighbors (ENN)~\citep{alejoenn}. 
Their method, evaluated on the Kaggle credit card dataset, demonstrates that this hybrid technique can significantly enhance the performance of ensemble models (including RF, KNN, and AdaBoost) and a voting ensemble.
By first oversampling the minority class and then using ENN to remove noisy samples, they achieve high scores in accuracy, precision, recall, F1, and AUC, outperforming many traditional oversamplers and showing that careful data balancing is crucial for robust fraud detection.
Overall, the prevailing trend in recent literature is to leverage generative models, typically trained only on minority class data, as advanced oversamplers.

\subsection{Gaps: Classifier-Guided Latent Shaping}
While the integration of auxiliary classifiers within generative frameworks has been explored to improve training stability and class-conditional sample quality \citep{sutedja2020imbalanced, engelmann2021conditional}, existing approaches primarily treat the classifier as a black-box supervisory signal. In auxiliary-classifier GANs, the classifier loss is typically used to encourage the generator to produce samples that are easily recognizable as belonging to a target class, without explicitly constraining the structure of the latent space or the geometry of the generated data manifold. As a result, these methods focus on class-conditional realism rather than on shaping representations that are discriminative with respect to downstream decision boundaries.
Moreover, prior GAN-based oversampling strategies are generally trained exclusively on minority-class data or generate samples conditioned only on the class label, which risks reinforcing local density without accounting for the global relationship between minority and majority classes. This can lead to synthetic samples that appear plausible in isolation but do not meaningfully improve class separation when evaluated against the full data distribution.
In contrast, our approach integrates an \textit{interpretable} classification head, the Causal Prototype Attention Classifier (CPAC), directly into the latent space of a VAE-GAN. Unlike standard auxiliary classifiers, CPAC does not merely provide a label-based loss to the generator. Instead, it actively structures the latent manifold through two explicit mechanisms: \textit{Prototype Anchoring}, which aligns class-specific prototypes with the centroids of latent clusters, and a \textit{Causal Attention} mechanism that assigns differential importance to latent dimensions based on their contribution to discrimination.
This design induces a bidirectional interaction between representation learning and sample generation: the generator is guided not only to produce class-consistent samples, but to position them in regions of the latent space that clarify the separation between fraud and non-fraud classes. By embedding discriminative structure directly into the generative process, the proposed VAE-GAN+CPAC framework goes beyond auxiliary-classifier supervision and yields synthetic samples that are both realistic and decision-aware, addressing a key limitation of prior classifier-guided GAN-based oversampling methods.

\section{Methodologies}
\label{sec:methodologies}
This section introduces the proposed CPAC model and evaluates its performance against standard classifiers. We also describe the proposed classifier-guided latent shaping strategy and analyze its effect on latent space structure and class separability. Then we will introduce how we used a classifier to influence and help the VAE-GAN latent space, shape a better representation of the two classes. Ultimately, it will be shown how the CPAC classifier might be a better fit for a classification head due to its nature and its structure and its results will be compared to other classification heads.

\subsection{Dataset and Preprocessing}
We conduct our experiments on the publicly available Credit Card Fraud Detection dataset, hosted on Kaggle~\citep{dal2015dataset}. This dataset, released by Worldline and the Machine Learning Group of ULB, contains 284,807 anonymized transactions made by European cardholders in September 2013. Only 492 of these are labeled as fraudulent, resulting in an extreme class imbalance (approximately 0.17\% fraud rate).
Each transaction includes 30 features, where 28 have been transformed via principal component analysis (PCA) to protect confidentiality. The remaining two features are the Time and Amount. The target variable Class is binary, with 1 indicating fraud and 0 otherwise.
To ensure each feature contributes uniformly to model training, we apply robust normalization to all input features. Specifically, for each feature \( x \), we compute its median \( \tilde{x} \) and interquartile range \( \text{IQR}(x) = Q_3(x) - Q_1(x) \), and normalize using Equation~\ref{eq:iqr}:
\begin{equation}
    \label{eq:iqr}
    \begin{aligned}
            x_{\text{norm}} = \frac{x - \tilde{x}}{\text{IQR}(x)}
    \end{aligned}
\end{equation}
This transformation centers features around zero and scales them while remaining robust to outliers, a crucial property in fraud detection where anomalies naturally exhibit extreme values. For features where \( \text{IQR}(x) = 0 \), we default to a unit divisor to prevent numerical instability.
We then split the dataset into 80\% training, 10\% validation, and 10\% test sets as shown in Table~\ref{tab:dataset-split}. Stratified sampling is used to preserve the class distribution across splits. Importantly, the splits are strictly disjoint: each transaction is assigned to exactly one subset, and no samples or derived information are shared across training, validation, and test sets, preventing any form of data leakage between partitions.
\begin{table}[!htp]
  \caption{Dataset split by class (80\% train, 10\% validation, 10\% test).}
  \label{tab:dataset-split}
  \centering
  \begin{tabular}{lccc}
    \toprule
    \textbf{Set}         & \textbf{Normal (0)} & \textbf{Fraud (1)} & \textbf{Total} \\
    \midrule
    Train (80\%)      & 227,452 & 394 & 227,846 \\
    Validation (10\%) & 28,431  & 49  & 28,480 \\
    Test (10\%)       & 28,432  & 49  & 28,481 \\
    \midrule
    Total             & 284,315 & 492 & 284,807 \\
    \bottomrule
  \end{tabular}
\end{table}

\subsection{Oversampling Strategies}

To address the extreme class imbalance in credit card fraud detection, we implemented and compared two oversampling strategies: SMOTE and a custom VAE–GAN pipeline. Both techniques were used to synthetically augment the minority (fraudulent) class, and the generated samples were added only to the training set, leaving the evaluation and test sets only with pure transactional data, emulating a real-life deploy scenario. We chose, 50, 75, 100 samples to generate for two main reasons: the first one is that using a higher number of samples could lead classifiers to overfit, especially if the number of generated frauds is higher than the original number in the dataset and the synthetic data is not of high quality. The second reason is that most of times, despite the increasing number of frauds in the training set, the models inevitably plateau as we will see in the next section.
\subsubsection{SMOTE-Based Oversampling}
\begin{figure}[!ht]
  \centering
  \includegraphics[width=1\linewidth]{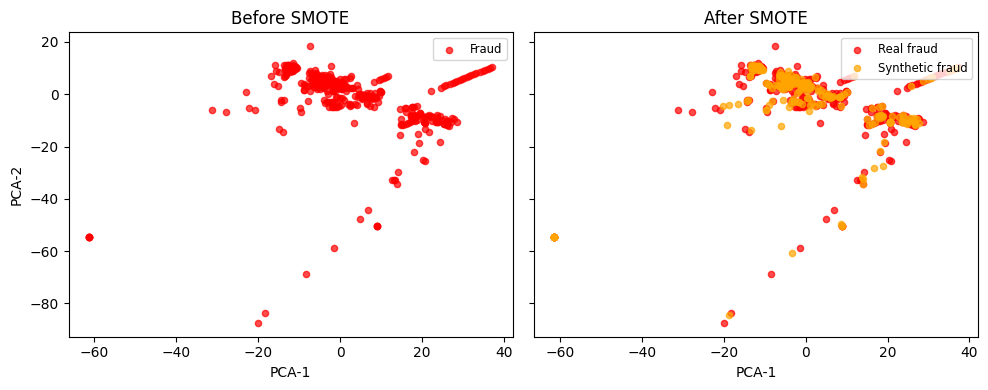}
  \caption{PCA plots comparing frauds distribution before and after SMOTE oversampling.}
  \label{fig:smote-ovs}
\end{figure}

The Synthetic Minority Over-sampling Technique (SMOTE) is a widely-used baseline for addressing class imbalance. Rather than learning the data distribution, SMOTE interpolates directly between existing minority-class samples. For two fraud instances $\mathbf{x}_i$ and $\mathbf{x}_j$, it generates a synthetic sample $\tilde{\mathbf{x}}$ along the line connecting them (Equation~\ref{eq:smote}):
\begin{equation}
    \label{eq:smote}
    \tilde{\mathbf{x}} = \mathbf{x}_i + \alpha \cdot (\mathbf{x}_j - \mathbf{x}_i), \qquad \alpha \sim \mathcal{U}(0, 1),
\end{equation}
where $\alpha$ is uniformly sampled. This process, repeated with each sample’s $k$ nearest neighbors, creates new minority points distributed across the feature space.
In our experiments, we generated 50, 75, and 100 synthetic frauds using SMOTE and merged them into the training set. While SMOTE is simple and effective, it can produce overly smooth or redundant samples, especially when the minority class has complex or non-linear structure. As shown in Figure~\ref{fig:smote-ovs}, SMOTE’s interpolated samples often “connect the dots” between real fraud clusters, potentially resulting in synthetic points that are too similar to the originals. Despite remaining a strong baseline, SMOTE can be outperformed by generative models that better capture the underlying data distribution in highly imbalanced settings.

\subsubsection{VAE–GAN Oversampling}
\begin{figure}[!ht]
  \centering
  \includegraphics[width=1\linewidth]{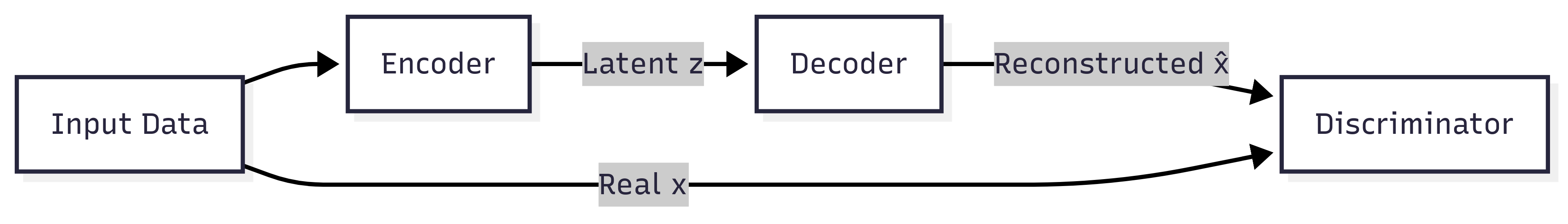}
  \caption{Diagram portraying the structure of the VAE-GAN.}
  \label{fig:vae-gan}
\end{figure}

\begin{figure}[!ht]
  \centering
  \includegraphics[width=1\linewidth]{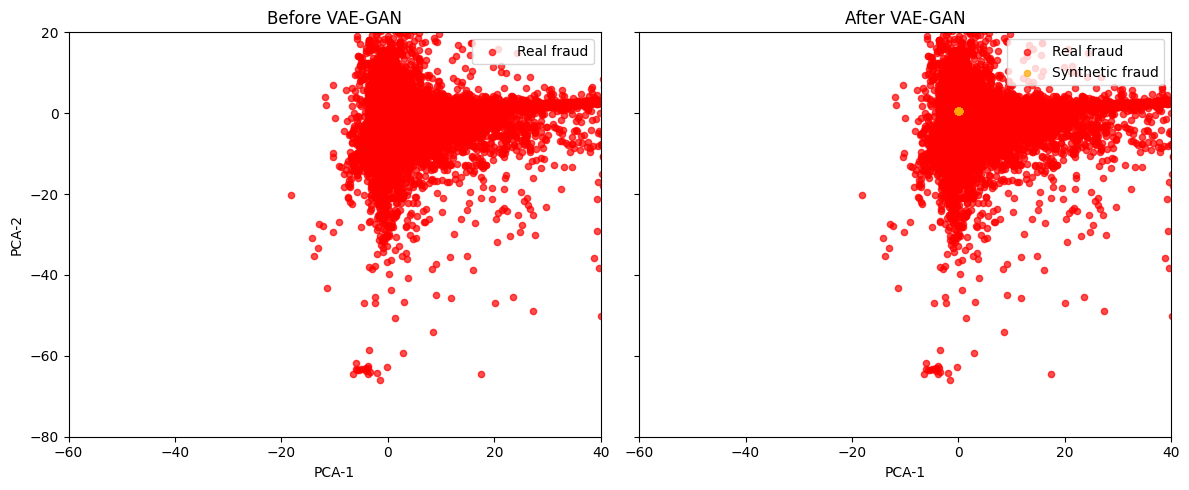}
  \caption{PCA plots comparing frauds distribution before and after VAE-GAN oversampling.}
  \label{fig:vagean-ovs}
\end{figure}

The Variational Autoencoder-Generative Adversarial Network (VAE-GAN) has emerged as a powerful approach for generating synthetic data in imbalanced classification problems (its structure is visible in Figure~\ref{fig:vae-gan}). In accordance with prevailing practice in fraud detection, we employ the VAE-GAN exclusively as a minority-class oversampler: it is trained using only genuine fraud transactions, then used to synthesize new fraud-like samples that supplement the training set.
The VAE-GAN consists of three neural modules: an encoder $E_\phi$, a decoder $D_\theta$, and a discriminator $C_\psi$. The encoder maps an input $\mathbf{x} \in \mathbb{R}^d$ through multiple hidden layers to the parameters of a multivariate Gaussian: mean vector $\boldsymbol{\mu}$ and log-variance $\log\boldsymbol{\sigma}^2$. Using the reparameterization trick, the latent code (Equation~\ref{eq:latent}) is sampled as
\begin{equation}
    \label{eq:latent}
    \mathbf{z} = \boldsymbol{\mu} + \boldsymbol{\sigma} \odot \boldsymbol{\epsilon}, \quad \boldsymbol{\epsilon} \sim \mathcal{N}(0, I).
\end{equation}
The decoder $D_\theta$ reconstructs the input from $\mathbf{z}$, while the discriminator $C_\psi$ distinguishes real from reconstructed samples.
Training proceeds by minimizing a weighted sum of three losses:
\begin{itemize}
    \item The \textbf{VAE loss} (Equation~\ref{eq:vae_loss}), which includes both a reconstruction error and a Kullback–Leibler divergence term:
    \begin{align}
        \label{eq:vae_loss}
           \mathcal{L}_{\mathrm{VAE}} = \mathbb{E}_{q_\phi(\mathbf{z}|\mathbf{x})}\!\left[\|\mathbf{x} - D_\theta(\mathbf{z})\|_2^2\right] 
        + \beta \cdot \mathrm{KL}\!\left(q_\phi(\mathbf{z}|\mathbf{x})\,\|\,p(\mathbf{z})\right), 
    \end{align}
    where $p(\mathbf{z})$ is the standard normal prior and $\beta$ controls the KL penalty.
    \item The \textbf{GAN loss} (Equation~\ref{eq:gan_loss}) for the discriminator, encouraging $C_\psi$ to distinguish real fraud samples from reconstructions:
    \begin{equation}
        \label{eq:gan_loss}
        \mathcal{L}_{\mathrm{GAN}} = -\mathbb{E}_{\mathbf{x} \sim p_{\mathrm{data}}}\!\left[\log C_\psi(\mathbf{x})\right]
        - \mathbb{E}_{\hat{\mathbf{x}} \sim D_\theta(\mathbf{z})}\!\left[\log(1 - C_\psi(\hat{\mathbf{x}}))\right].
    \end{equation}
    \item The \textbf{generator adversarial loss} (Equation~\ref{eq:adv_loss}), which pushes the decoder to generate samples that the discriminator cannot distinguish from real frauds:
    \begin{equation}
        \label{eq:adv_loss}
        \mathcal{L}_{\mathrm{Adv}} = -\mathbb{E}_{\hat{\mathbf{x}} \sim D_\theta(\mathbf{z})}\!\left[\log C_\psi(\hat{\mathbf{x}})\right].
    \end{equation}
\end{itemize}
During training, the encoder and decoder are optimized together to minimize both reconstruction and adversarial losses, while the discriminator is trained to distinguish real from generated samples. Early stopping and validation are used to ensure generalization. After training, the decoder generates new fraud samples by sampling from the learned latent distribution, augmenting the training set for subsequent classification. This VAE-GAN-based oversampling is widely used for its ability to produce more realistic and varied fraud examples than simpler interpolation techniques like SMOTE. In our experiments, we generated and merged 50, 75 and 100 synthetic fraud samples into the training data. However, as illustrated in Figure~\ref{fig:vagean-ovs}, such oversampling often concentrates the synthetic data in a narrow region of latent space, which can make downstream classifiers either overconfident or poorly calibrated, and limits cluster separation. This limitation provides the motivation for our classifier-guided methods.

\subsubsection{SMOTE vs VAE-GAN Oversampling}
The fundamental difference between SMOTE and VAE–GAN oversampling lies in how synthetic minority-class examples are generated and distributed. SMOTE creates new samples by interpolating between real frauds and their nearest neighbors, filling gaps and uniformly spreading synthetic points within known clusters. While easy to use and computationally light, SMOTE may also generate borderline samples that overlap with the majority class. In contrast, VAE–GAN learns a latent representation and generates new frauds by sampling from a global latent prior, which often leads to denser clustering around the main fraud mode but can miss peripheral fraud patterns. VAE–GAN models require more careful design and tuning, but can create more realistic, nonlinear samples that better reflect complex feature relationships. Therefore, SMOTE is suited for quick and broad coverage of known minority regions, whereas VAE–GAN offers more expressive synthetic data at the cost of increased complexity and training effort. The choice ultimately depends on the application's requirements and available resources.

\subsection{Baseline Classifiers}

To benchmark the performance of our generative and prototype-based approaches, we evaluated three standard classifiers widely used in fraud detection: Logistic Regression, Random Forest, KNN and XGBoost. These models provide strong baselines due to their interpretability, ensemble nature, and ability to handle imbalanced data with appropriate modifications. To assess the performance of all classifiers in this work, we report four standard metrics: precision, recall, F1-score, and AUC-ROC. Each metric captures a different aspect of performance for highly imbalanced classification problems such as fraud detection.

\subsection{Causal Prototype Attention Classifier (CPAC)}
The Causal Prototype Attention Classifier (CPAC) embeds interpretable, class‐aware structure directly into a lightweight neural module. Given an input $\mathbf{x}\in\mathbb{R}^d$, CPAC learns two prototype vectors (Equation~\ref{eq:protss})

\begin{equation}
    \label{eq:protss}
    \begin{aligned}
        \mathbf{p}_0,\;\mathbf{p}_1\;\in\;\mathbb{R}^d
    \end{aligned}
\end{equation}

representing the centroids of the non‐fraud and fraud classes. An attention network (Equation~\ref{eq:att})

\begin{equation}
    \label{eq:att}
    \begin{aligned}
        \mathbf{w} \;=\; \mathrm{Att}(\mathbf{x}) \;=\;\sigma\bigl(W_2\,\mathrm{ReLU}(W_1\mathbf{x}+b_1)+b_2\bigr)
        \;\;\in(0,1)^d 
    \end{aligned}
\end{equation}

where $\mathbf{x} \in \mathbb{R}^d$ is the latent input, $W_1$ and $W_2$ are learnable weight matrices, $b_1$ and $b_2$ are biases, $\mathrm{ReLU}$ is the rectified linear unit, and $\sigma$ is the sigmoid function. The output $\mathbf{w} \in (0,1)^d$ is a per-feature attention mask. It produces a per‐feature mask highlighting dimensions most predictive of fraud.  A learnable scale $\alpha>0$ adjusts sensitivity, and we compute weighted squared distances (Equation~\ref{eq:weight_dis}).

\begin{equation}
    \label{eq:weight_dis}
    \begin{aligned}
            d_c(\mathbf{x}) \;=\; \alpha\,\sum_{i=1}^d w_i\,\bigl(x_i - p_{c,i}\bigr)^2
            \quad,\;c\in\{0,1\}.
    \end{aligned}
\end{equation}

where $w_i$ is the $i$-th element of the attention vector $\mathbf{w}$, $x_i$ is the $i$-th feature of the latent vector $\mathbf{x}$, $p_{c,i}$ is the $i$-th coordinate of the prototype for class $c$ ($c=0$ for non-fraud, $c=1$ for fraud), $\alpha$ is a learnable scaling factor, and $d$ is the latent dimensionality.

Interpreting negative distances as logits (Equation~\ref{eq:neg_dis}),

\begin{equation}
    \label{eq:neg_dis}
    \begin{aligned}
        \boldsymbol{\ell}(\mathbf{x})
        =\begin{bmatrix}-d_0(\mathbf{x})\\-d_1(\mathbf{x})\end{bmatrix}
        \quad\Longrightarrow\quad
        \hat y \;=\;\mathrm{softmax}\bigl(\boldsymbol{\ell}(\mathbf{x})\bigr)_1,
    \end{aligned}
\end{equation}

where $d_0(\mathbf{x})$ and $d_1(\mathbf{x})$ are the weighted distances to the non-fraud and fraud prototypes, respectively, $\boldsymbol{\ell}(\mathbf{x})$ is the vector of negative distances used as logits, and $\mathrm{softmax}(\cdot)_1$ denotes the softmax probability for class 1 (fraud). This yields the predicted fraud probability $\hat y\in(0,1)$.

\begin{figure}[!ht]
  \centering
  \includegraphics[width=1\linewidth]{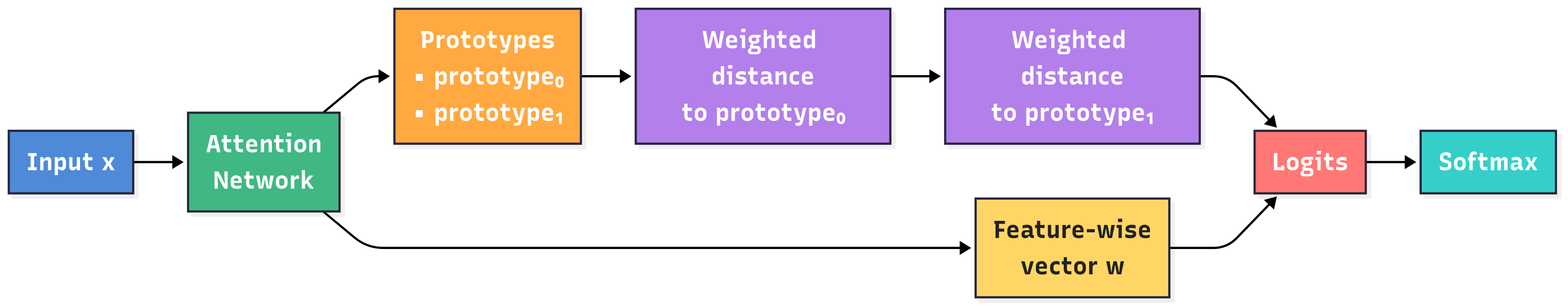}
  \caption{Architecture of the CPAC model. Each input is compared to class prototypes using an attention-weighted distance, followed by softmax scoring.}
  \label{fig:cpac-diagram}
\end{figure}

The term “causal” here is used loosely to suggest that the attention weights may help highlight which latent features have a higher impact on the outcome. Its structure can be visualised in Figure~\ref{fig:cpac-diagram}.

\subsubsection{Training Loss}  
To counter the extreme imbalance, we adopt the Focal Loss~\citep{lin2017focal}, which adds two hyperparameters (Equation~\ref{eq:focal}), \(\alpha_{\mathrm{FL}}\) and \(\gamma\) to the standard binary cross‐entropy:

\begin{align}
\mathcal{L}_{\mathrm{FL}}(y,\hat y)
  =\; & -\,\alpha_{\mathrm{FL}}\,(1-\hat y)^{\gamma}\,y\,\log\hat y \notag \\
      & -\; (1-\alpha_{\mathrm{FL}})\,\hat y^{\gamma}\,(1-y)\,\log(1-\hat y)
\label{eq:focal}
\end{align}

Where:
\begin{itemize}
  \item \(\alpha_{\mathrm{FL}}\in[0,1]\) balances the importance of the two classes.  
        We set \(\alpha_{\mathrm{FL}}=0.95\) to give more weight to the minority (fraud) class.
  \item \(\gamma\ge0\) is the focusing parameter.  When \(\gamma=0\), \(\mathcal{L}_{\mathrm{FL}}\) reduces to ordinary cross‐entropy.  
        As \(\gamma\) grows, well-classified examples (where \(\hat y\) is close to the true label) incur much smaller loss, 
        forcing the model to concentrate on harder, often minority‐class cases.  
        We performed a grid search for the most optimal values for \(\alpha_{\mathrm{FL}}\) and \(\gamma\) and we found out that the best results are obtained with 0.95 and 2.0 respectively.
\end{itemize}

By tuning \(\alpha_{\mathrm{FL}}\) and \(\gamma\), Focal Loss both re‐weights the underrepresented class and focuses learning on its most difficult examples (Alg.~\ref{alg:cpac_training}), which is crucial in fraud detection.
Model selection uses a composite score (Equation~\ref{eq:comp_score}):

\begin{equation}
    \label{eq:comp_score}
    \begin{aligned}
       S \;=\; 0.50\,\mathrm{Precision}\;+\;0.50\,\mathrm{Recall} 
    \end{aligned}
\end{equation}

on the validation set. We checkpoint the CPAC weights whenever $S$ improves, and invoke early‐stopping after $p$ epochs without gain.

\begin{algorithm}[!ht]
\caption{Training of the Causal Prototype Attention Classifier}
\label{alg:cpac_training}
\footnotesize
\begin{algorithmic}[1]
\REQUIRE Training set $\mathcal{D}_{\mathrm{train}} = \{(\mathbf{x}_i, y_i)\}$
\REQUIRE Validation set $\mathcal{D}_{\mathrm{val}}$
\REQUIRE Initial parameters $\Theta = \{\mathbf{p}_0, \mathbf{p}_1, W_1, W_2, b_1, b_2, \alpha\}$
\REQUIRE Focal Loss parameters $\alpha_{\mathrm{FL}}, \gamma$
\REQUIRE Patience $p$
\STATE $S_{\mathrm{best}} \leftarrow 0$
\FOR{each epoch}
    \FOR{each mini-batch $(\mathbf{x}, y) \subset \mathcal{D}_{\mathrm{train}}$}
        \STATE $\hat y \leftarrow \mathrm{CPAC}(\mathbf{x}; \Theta)$ 
        \STATE Compute Focal Loss $\mathcal{L}_{\mathrm{FL}}(y, \hat y)$
        \STATE Update $\Theta \leftarrow \Theta - \eta \nabla_\Theta \mathcal{L}_{\mathrm{FL}}$
    \ENDFOR
    \STATE Evaluate Precision and Recall on $\mathcal{D}_{\mathrm{val}}$
    \STATE $S \leftarrow 0.5\,\mathrm{Precision} + 0.5\,\mathrm{Recall}$
    \IF{$S > S_{\mathrm{best}}$}
        \STATE Save model checkpoint
        \STATE $S_{\mathrm{best}} \leftarrow S$
        \STATE Reset early-stopping counter
    \ELSE
        \STATE Increment early-stopping counter
    \ENDIF
    \IF{early-stopping counter $\ge p$}
        \STATE \textbf{break}
    \ENDIF
\ENDFOR
\end{algorithmic}
\end{algorithm}

\subsection{Adaptive Threshold Selection via Differentiable Agent}
To improve classification under class imbalance, we introduce a differentiable agent that adaptively learns the optimal classification threshold \(\tau^*\) for maximizing the F1 score. Instead of using the default \(\tau = 0.5\), we parameterize the threshold as a scalar \(\theta\), mapping it to \(\tau = \sigma(\theta)\) (where \(\sigma\) is the sigmoid function) so that \(\tau\) lies in (0,1).
For each sample \(i\), with predicted probability \(p_i\) and ground truth \(y_i \in \{0,1\}\), the agent approximates a hard threshold with a smooth sigmoid function (Equation~\ref{eq:bin_dec}):

\begin{equation}
    \label{eq:bin_dec}
    \begin{aligned}
        \hat{y}_i = \sigma\bigl(\beta(p_i - \tau)\bigr),
    \end{aligned}
\end{equation}

where $\hat{y}_i$ is the soft binary prediction for sample $i$, $p_i$ is the predicted probability, $\tau$ is the learned threshold, and $\beta$ controls the sharpness of the sigmoid (making the function step-like).
The loss for optimizing the threshold is given by Equation~\ref{eq:dummy_loss}:

\begin{equation}
    \label{eq:dummy_loss}
    \begin{aligned}
        \mathcal{L} = \frac{1}{N} \sum_{i=1}^N \left(\hat{y}_i - y_i\right)^2,
    \end{aligned}
\end{equation}

where $N$ is the batch size, $\hat{y}_i$ is the soft prediction, and $y_i$ is the true binary label.
By minimizing this loss via gradient descent and monitoring validation F1, the agent effectively learns an optimal threshold suited to the classifier’s output probabilities. This method consistently yielded improved F1 scores, often with a learned threshold above 0.7, especially when paired with classifiers such as CPAC or those using VAE–GAN encoders.

\subsection{VAE-GAN with Classification Heads}
Most state-of-the-art techniques~\citep{Wang2022, Ding2023, Shi2025_balVAE}, train the generative model on only the fraud (minority) class. This classical approach has a crucial limitation: it fails to expose the model to the characteristics of non-fraudulent (majority) data, resulting in a generator that can merely interpolate among known frauds, often producing synthetic samples that are near-duplicates, lacking true discriminative power. The latent space learned in such a setup is inevitably narrow and uninformative about what actually distinguishes fraud from normality.
In contrast, our approach is fundamentally different. We jointly train the VAE-GAN with a classification head on the full dataset, even though the generative (VAE-GAN) component is optimized only on the minority class. The critical distinction is that the encoder, which is shared between the generator and the classifier head, receives supervised feedback from both classes via the classification objective. This means that the latent space, as learned by the encoder, encodes information about the entire data distribution,both fraud and non-fraud.
This approach fundamentally differs from conventional pipelines in which the generative model is trained exclusively on the minority class. The VAE-GAN and classifier pipeline is fully aware of both classes, since the encoder’s parameters are influenced by the generative loss applied to fraud samples and the classification loss computed over the entire dataset. Focusing the generative loss on the minority class is a deliberate choice that enables the decoder to specialize in high-fidelity, targeted synthesis of frauds. At the same time, the classifier head continuously regularizes the encoder, ensuring that it organizes the latent space to distinguish between both fraud and non-fraud classes. As a result, the latent space does not lose information about the normal class; rather, it is shaped to maximize class discrimination. The encoder integrates the objectives of both components, while the decoder utilizes this enriched representation to generate synthetic frauds that are not only realistic but also truly distinct within the broader data context. Thus, although the generative focus is on the minority class, the overall supervision ensures that the synthetic samples are robust, generalizable, and valuable for distinguishing between classes.

\subsubsection{Classifier Heads (MLP Variants)}
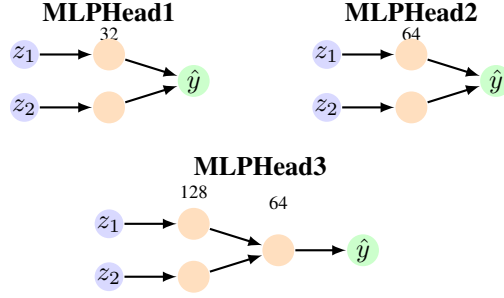
\begin{figure}[ht]
\centering
\begin{tikzpicture}[x=1.6cm, y=1.4cm, >=latex, thick]

\node[draw=none, fill=none] (h1title) at (0.7,2.0) {\textbf{MLPHead1}};
\node[draw=none, fill=none] (h1units) at (0.7,1.8) {\scriptsize 32};
\node[fill=blue!15,circle,minimum size=10pt,inner sep=0pt] (z1a) at (0,1.6) {$z_1$};
\node[fill=blue!15,circle,minimum size=10pt,inner sep=0pt] (z2a) at (0,1.1) {$z_2$};
\node[fill=orange!25,circle,minimum size=12pt,inner sep=0pt] (h1n1) at (0.7,1.6) {};
\node[fill=orange!25,circle,minimum size=12pt,inner sep=0pt] (h1n2) at (0.7,1.1) {};
\node[fill=green!20,circle,minimum size=12pt,inner sep=0pt] (h1out) at (1.4,1.35) {$\hat{y}$};
\draw[->] (z1a) -- (h1n1);
\draw[->] (z2a) -- (h1n2);
\draw[->] (h1n1) -- (h1out);
\draw[->] (h1n2) -- (h1out);

\node[draw=none, fill=none] (h2title) at (3.2,2.0) {\textbf{MLPHead2}};
\node[draw=none, fill=none] (h2units) at (3.2,1.8) {\scriptsize 64};
\node[fill=blue!15,circle,minimum size=10pt,inner sep=0pt] (z1b) at (2.5,1.6) {$z_1$};
\node[fill=blue!15,circle,minimum size=10pt,inner sep=0pt] (z2b) at (2.5,1.1) {$z_2$};
\node[fill=orange!25,circle,minimum size=12pt,inner sep=0pt] (h2n1) at (3.2,1.6) {};
\node[fill=orange!25,circle,minimum size=12pt,inner sep=0pt] (h2n2) at (3.2,1.1) {};
\node[fill=green!20,circle,minimum size=12pt,inner sep=0pt] (h2out) at (3.9,1.35) {$\hat{y}$};
\draw[->] (z1b) -- (h2n1);
\draw[->] (z2b) -- (h2n2);
\draw[->] (h2n1) -- (h2out);
\draw[->] (h2n2) -- (h2out);

\node[draw=none, fill=none] (h3title) at (1.95,0.55) {\textbf{MLPHead3}};
\node[draw=none, fill=none] (h3units1) at (1.4,0.30) {\scriptsize 128};
\node[draw=none, fill=none] (h3units2) at (2.1,0.20) {\scriptsize 64};
\node[fill=blue!15,circle,minimum size=10pt,inner sep=0pt] (z1c) at (0.7,0) {$z_1$};
\node[fill=blue!15,circle,minimum size=10pt,inner sep=0pt] (z2c) at (0.7,-0.5) {$z_2$};
\node[fill=orange!25,circle,minimum size=12pt,inner sep=0pt] (h3n1) at (1.4,0) {};
\node[fill=orange!25,circle,minimum size=12pt,inner sep=0pt] (h3n2) at (1.4,-0.5) {};
\node[fill=orange!25,circle,minimum size=12pt,inner sep=0pt] (h3n3) at (2.1,-0.25) {};
\node[fill=green!20,circle,minimum size=12pt,inner sep=0pt] (h3out) at (2.8,-0.25) {$\hat{y}$};
\draw[->] (z1c) -- (h3n1);
\draw[->] (z2c) -- (h3n2);
\draw[->] (h3n1) -- (h3n3);
\draw[->] (h3n2) -- (h3n3);
\draw[->] (h3n3) -- (h3out);

\end{tikzpicture}
\caption{Architectures of the three MLP heads used for classification on the 2-dimensional latent space. Hidden units are indicated above each hidden layer.}
\label{fig:mlp_heads}
\end{figure}

We consider three multilayer perceptron (MLP) heads, as shown in Figure~\ref{fig:mlp_heads}, of increasing complexity, each implementing a function $h_\theta: \mathbb{R}^d \to [0,1]$, with $d=2$ in our experiments:

\begin{enumerate}
    \item \textbf{MLPHead1}: One hidden layer with 32 units (ReLU), output through a sigmoid (Equation~\ref{eq:mlp1}):
    \begin{equation}
        \label{eq:mlp1}
        \begin{aligned}
            h_1(z) = \sigma \left( W_2\, \mathrm{ReLU}(W_1 z + b_1) + b_2 \right)
        \end{aligned}
    \end{equation}
    where $W_1 \in \mathbb{R}^{32 \times 2}$, $b_1 \in \mathbb{R}^{32}$, $W_2 \in \mathbb{R}^{1 \times 32}$, $b_2 \in \mathbb{R}$.

    \item \textbf{MLPHead2}: One hidden layer with 64 units, batch normalization, dropout ($p=0.2$), ReLU and sigmoid (Equation~\ref{eq:mlp2}):
    \begin{equation}
        \label{eq:mlp2}
        \begin{aligned}
            h_2(z) = \sigma \left( W_2\, \mathrm{Dropout}\big(\mathrm{ReLU}(\mathrm{BN}(W_1 z + b_1))\big) + b_2 \right)
        \end{aligned}
    \end{equation}
    with $W_1 \in \mathbb{R}^{64 \times 2}$, $b_1 \in \mathbb{R}^{64}$, $W_2 \in \mathbb{R}^{1 \times 64}$, $b_2 \in \mathbb{R}$.

    \item \textbf{MLPHead3}: Two hidden layers with 128 and 64 units (ReLU), output through a sigmoid (Equation~\ref{eq:mlp3}):
    \begin{equation}
        \label{eq:mlp3}
        \begin{aligned}
            h_3(z) = \sigma \left( W_3\, \mathrm{ReLU}(W_2\, \mathrm{ReLU}(W_1 z + b_1) + b_2) + b_3 \right)
        \end{aligned}
    \end{equation}
    where $W_1 \in \mathbb{R}^{128 \times 2}$, $b_1 \in \mathbb{R}^{128}$, $W_2 \in \mathbb{R}^{64 \times 128}$, $b_2 \in \mathbb{R}^{64}$, $W_3 \in \mathbb{R}^{1 \times 64}$, $b_3 \in \mathbb{R}$.
\end{enumerate}

These structures will all be tested and evaluated as potential heads for our encoder.

\subsubsection{Joint Training Procedure}
The model is trained in two coordinated phases at each epoch:

\begin{enumerate}
    \item \textbf{VAE-GAN update}: For each mini-batch (from all classes), the encoder maps $x$ to $z, \mu, \log\sigma^2$, and the decoder reconstructs $x_{\text{rec}}$. The discriminator receives both $x$ and $x_{\text{rec}}$ and tries to distinguish real from generated data. The VAE-GAN is trained with the following loss (Equation~\ref{eq:vaegan_mlp}):
    \begin{equation}
        \label{eq:vaegan_mlp}
        \begin{aligned}
            \mathcal{L}_\mathrm{VAE-GAN} = \mathcal{L}_\mathrm{recon} + \mathcal{L}_\mathrm{KL} + \mathcal{L}_\mathrm{GAN}
        \end{aligned}
    \end{equation}
    where $\mathcal{L}_\mathrm{recon}$ is MSE between $x$ and $x_{\text{rec}}$, $\mathcal{L}_\mathrm{KL}$ is the KL divergence, and $\mathcal{L}_\mathrm{GAN}$ is the adversarial loss for the generator decoder.
    
    \item \textbf{Classifier head update}: Using the same mini-batch, the encoder's mean $\mu$ is passed to the classifier head $h_\theta$ to predict the label $y$. The head is trained with binary cross-entropy (Equation~\ref{eq:bce}):
    \begin{equation}
        \label{eq:bce}
        \begin{aligned}
            \mathcal{L}_\mathrm{clf}(y, \hat{y}) = -y \log \hat{y} - (1-y)\log(1-\hat{y}), \quad \hat{y} = h_\theta(\mu)
        \end{aligned}
    \end{equation}
    The gradient of $\mathcal{L}_\mathrm{clf}$ flows back not only to the classifier head parameters $\theta$, but also to the encoder parameters. As a result, the encoder is explicitly encouraged to organize the latent means $\mu$ so that different classes become more separable for the downstream classifier. This joint training guides the encoder to learn a latent representation where fraud and non-fraud samples are more easily discriminated.
\end{enumerate}

\begin{figure}[!ht]
    \centering
    \begin{subfigure}[b]{0.30\textwidth}
        \centering
        \includegraphics[width=\textwidth]{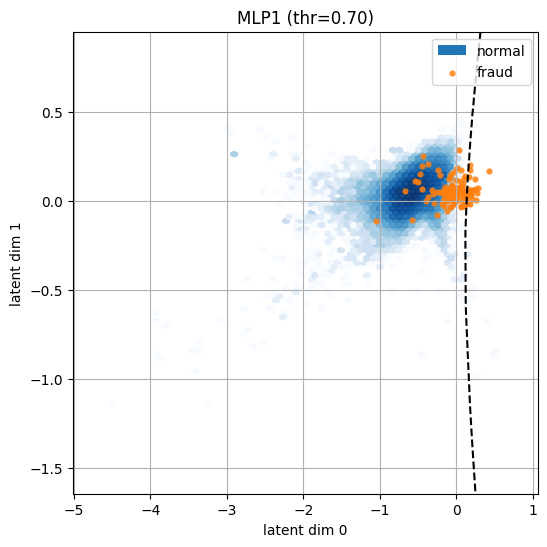}
        \caption{MLPHead1}
        \label{fig:mlp1}
    \end{subfigure}
    \hfill
    \begin{subfigure}[b]{0.30\textwidth}
        \centering
        \includegraphics[width=\textwidth]{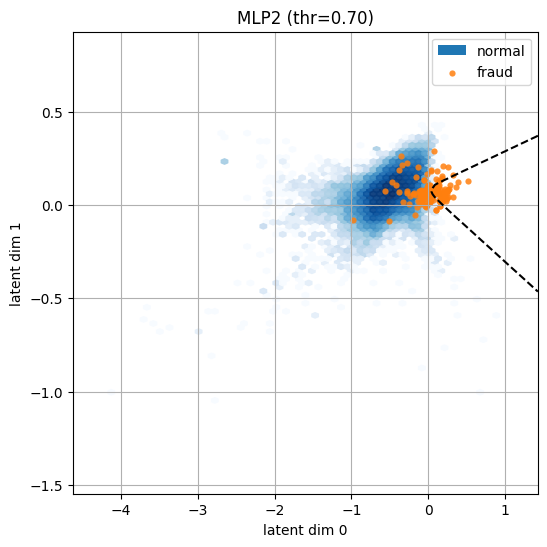}
        \caption{MLPHead2}
        \label{fig:mlp2}
    \end{subfigure}
    \hfill
    \begin{subfigure}[b]{0.30\textwidth}
        \centering
        \includegraphics[width=\textwidth]{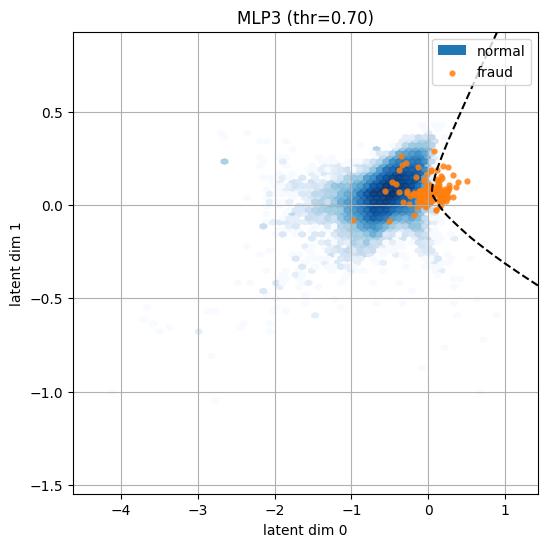}
        \caption{MLPHead3}
        \label{fig:mlp3}
    \end{subfigure}
    \caption{PCA plots showing cluster separation in the latent space for different MLP heads. \color{black}Lower density points are less colored, while higher density (bigger overlap) are more highlighted.\color{black}}
    \label{fig:mlp_heads1}
\end{figure}

The pseudo-code of the algorithm can be observed in Alg.~\ref{alg:mlp_joint_training}.
Despite all three experiments with all the MLPs (as shown in Figures~\ref{fig:mlp1}, \ref{fig:mlp2}, \ref{fig:mlp3}) clearly suggest at a slight cluster separation, the overlap between the two classes is still significant. The first and third MLPs produces similar results despite the different structures while the second MLP is the only one leaning into a polynomial decision boundary that could help separating the two clusters. These results prove that we need a more powerful network capable of adapting into the encoder logic and working on a non-linear and more complex space. Due to the substantial overlap between the two classes, using this latent representation to generate additional fraud samples degrades the performance of downstream classifiers.

\begin{algorithm}[!ht]
\caption{Joint Training of VAE-GAN with Classifier Head}
\label{alg:mlp_joint_training}
\footnotesize
\begin{algorithmic}[1]
\REQUIRE Dataset $\mathcal{D} = \{(x_i, y_i)\}$, encoder $E_\phi$, decoder $G_\psi$, discriminator $D_\omega$, classifier head $h_\theta$
\REQUIRE Learning rates $\eta_\phi, \eta_\psi, \eta_\omega, \eta_\theta$
\FOR{each epoch}
    \FOR{each mini-batch $\{(x, y)\} \subset \mathcal{D}$}
        \STATE \textbf{Encoder forward:} $(\mu, \log \sigma^2) \leftarrow E_\phi(x)$
        \STATE Sample latent code $z \sim \mathcal{N}(\mu, \sigma^2)$
        \STATE \textbf{Decoder forward:} $x_{\text{rec}} \leftarrow G_\psi(z)$
        \STATE \textbf{Discriminator update:}
        \STATE \hspace{0.5cm} Compute $\mathcal{L}_{\mathrm{GAN}}(D_\omega(x), D_\omega(x_{\text{rec}}))$
        \STATE \hspace{0.5cm} Update $\omega \leftarrow \omega - \eta_\omega \nabla_\omega \mathcal{L}_{\mathrm{GAN}}$
        \STATE \textbf{VAE-GAN loss:}
        \STATE \hspace{0.5cm} $\mathcal{L}_{\mathrm{VAE-GAN}} = \mathcal{L}_{\mathrm{recon}}(x, x_{\text{rec}}) + \mathcal{L}_{\mathrm{KL}}(\mu, \log \sigma^2) + \mathcal{L}_{\mathrm{GAN}}$
        \STATE Update $\phi, \psi \leftarrow \phi, \psi - \eta_{\phi,\psi} \nabla_{\phi,\psi} \mathcal{L}_{\mathrm{VAE-GAN}}$
        \STATE \textbf{Classifier head update:}
        \STATE \hspace{0.5cm} $\hat{y} \leftarrow h_\theta(\mu)$
        \STATE \hspace{0.5cm} $\mathcal{L}_{\mathrm{clf}} = -y \log \hat{y} - (1-y)\log(1-\hat{y})$
        \STATE Update $\theta, \phi \leftarrow \theta, \phi - \eta_{\theta,\phi} \nabla_{\theta,\phi} \mathcal{L}_{\mathrm{clf}}$
    \ENDFOR
\ENDFOR
\end{algorithmic}
\color{black}
\end{algorithm}

\subsection{VAE-GAN with CPAC Head}
Similarly to how we did with the MLPs, we tested how the CPAC behaves as a classification head paired with the encoder. With the MLPs, there was a slight tendency to inner cluster separation in the encoder but the overlap was still significant. CPAC's structure, is very akin with the encoder, because its prototypes might offer an anchor point where the centroid of each cluster might find its position pushing for a better separation. The training is similar to what has already been done with the MLPs.

\subsubsection{Joint Training Procedure} 
At each training epoch, we alternate between:

\begin{enumerate}
    \item \textbf{VAE-GAN training:} For each batch, the encoder $E_\phi$ encodes $x$ to $(z, \mu, \log\sigma^2)$, the decoder $G_\psi$ reconstructs $x_\text{rec} = G_\psi(z)$, and the discriminator $D_\omega$ distinguishes $x$ from $x_\text{rec}$. The standard VAE-GAN loss (Equation~\ref{eq:vaegan_cpac}) is:
    \begin{equation}
        \label{eq:vaegan_cpac}
        \begin{aligned}
            \mathcal{L}_\text{VAE-GAN} = \mathcal{L}_\text{recon}(x, x_\text{rec}) + \mathcal{L}_\text{KL}(\mu, \log\sigma^2) + \mathcal{L}_\text{GAN}(D_\omega(x_\text{rec}), 1)
        \end{aligned}
    \end{equation}
    where $\mathcal{L}_\text{recon}$ is mean squared error (MSE), $\mathcal{L}_\text{KL}$ is the KL divergence, and $\mathcal{L}_\text{GAN}$ is the adversarial loss for the generator.

    \item \textbf{CPAC update:} In the same batch, the encoder’s mean $\mu$ is passed to the CPAC head, which computes distances (Equation~\ref{eq:dist}) to two learnable prototypes ($\mathbf{p}_0$, $\mathbf{p}_1$) via feature-wise attention weights $\mathbf{w}$:
    \begin{equation}
        \label{eq:dist}
        \begin{aligned}
            d_c = \alpha \sum_{j=1}^d w_j \left(\mu_j - (\mathbf{p}_c)_j\right)^2, \qquad c \in \{0, 1\}
        \end{aligned}
    \end{equation}
    where $w_j \in (0,1)$ are attention weights (from a neural branch), and $\alpha$ is a learnable scaling parameter. Fraud probability is given by Equation~\ref{eq:fr_probs}:
    \begin{equation}
        \label{eq:fr_probs}
        \begin{aligned}
            \hat{y} = \mathrm{Softmax}(-d_0, -d_1)   
        \end{aligned}
    \end{equation}
    The CPAC is trained to minimize the BCE loss.
    To further regularize learning, two penalties are added:
    \begin{itemize}
        \item \textbf{Scale penalty:} encourages the attention scaling parameter to stay bounded (Equation~\ref{eq:scale}):
        \begin{equation}
            \label{eq:scale}
            \begin{aligned}
                \mathcal{L}_\text{scale} = \lambda_\text{scale} \cdot \|\alpha\|^2
            \end{aligned}
        \end{equation}
        \item \textbf{Prototype anchoring:} aligns each prototype to the centroid of its class in latent space, encouraging the encoder to cluster samples around the correct prototype (Equation~\ref{eq:proto}):
        \begin{equation}
            \label{eq:proto}
            \begin{aligned}
                \mathcal{L}_\text{anchor} = \lambda_\text{anchor} \left( \|\mathbf{p}_0 - \bar{\mu}_0\|^2 + \|\mathbf{p}_1 - \bar{\mu}_1\|^2 \right)
            \end{aligned}
        \end{equation}
        where $\bar{\mu}_0$, $\bar{\mu}_1$ are the means of latent vectors in the current batch with $y=0$ and $y=1$ respectively.
    \end{itemize}
\end{enumerate}

The total loss (Equation~\ref{eq:tot_loss}) for CPAC becomes:
\begin{equation}
    \label{eq:tot_loss}
    \begin{aligned}
        \mathcal{L}_\text{CPAC-total} = \mathcal{L}_\text{clf} + \mathcal{L}_\text{scale} + \mathcal{L}_\text{anchor}
    \end{aligned}
\end{equation}
and the overall optimization alternates VAE-GAN and CPAC updates. Just like we did for the Focal Loss we performed a grid-search and found the best \(\lambda_\text{scale}\) and \(\lambda_\text{anchor}\) to be respectively 0.001 and 0.01. When using CPAC as a classification head within our VAE-GAN architecture, we switch to BCE loss. Just like the MLPs, the gradient of the CPAC loss gets backpropagated to both the CPAC parameters and also to the encoder parameter. After each epoch, we evaluate the CPAC on a held-out validation set. Early stopping is applied based on recall, conditional on maintaining a minimum precision, as in other head experiments.
The full joint optimization procedure of the VAE-GAN with CPAC head is summarized in Algorithm~\ref{alg:vaegan_cpac}.

\begin{algorithm}[!htp]
\caption{Joint Training of VAE-GAN with CPAC Head}
\label{alg:vaegan_cpac}
\footnotesize
\begin{algorithmic}[1]
\REQUIRE Training set $\mathcal{D}_{\mathrm{train}} = \{(x_i, y_i)\}$
\REQUIRE Validation set $\mathcal{D}_{\mathrm{val}}$
\REQUIRE Encoder $E_\phi$, decoder $G_\psi$, discriminator $D_\omega$
\REQUIRE CPAC parameters $\Theta_{\mathrm{CPAC}} = \{\mathbf{p}_0, \mathbf{p}_1, W_1, W_2, b_1, b_2, \alpha\}$
\REQUIRE Regularization weights $\lambda_{\mathrm{scale}}, \lambda_{\mathrm{anchor}}$
\REQUIRE Patience $p$
\STATE Initialize best validation score $S_{\mathrm{best}} \leftarrow 0$
\FOR{each epoch}
    \FOR{each mini-batch $(x, y) \subset \mathcal{D}_{\mathrm{train}}$}
        \STATE \textbf{Encoder forward:} $(\mu, \log \sigma^2) \leftarrow E_\phi(x)$
        \STATE Sample $z \sim \mathcal{N}(\mu, \sigma^2)$
        \STATE \textbf{Decoder forward:} $x_{\mathrm{rec}} \leftarrow G_\psi(z)$
        \STATE \textbf{Discriminator update:}
        \STATE \hspace{0.5cm} Compute $\mathcal{L}_{\mathrm{GAN}}(D_\omega(x_{\mathrm{rec}}), 1)$
        \STATE \hspace{0.5cm} Update $\omega \leftarrow \omega - \eta_\omega \nabla_\omega \mathcal{L}_{\mathrm{GAN}}$
        \STATE \textbf{VAE-GAN loss:}
        \STATE \hspace{0.5cm} $\mathcal{L}_{\mathrm{VAE-GAN}} = \mathcal{L}_{\mathrm{recon}} + \mathcal{L}_{\mathrm{KL}} + \mathcal{L}_{\mathrm{GAN}}$
        \STATE Update $\phi, \psi \leftarrow \phi, \psi - \eta_{\phi,\psi} \nabla_{\phi,\psi} \mathcal{L}_{\mathrm{VAE-GAN}}$
        \STATE \textbf{CPAC forward:}
        \STATE \hspace{0.5cm} $\mathbf{w} \leftarrow \mathrm{Att}(\mu)$
        \FOR{$c \in \{0,1\}$}
            \STATE \hspace{0.5cm} $d_c \leftarrow \alpha \sum_{j=1}^d w_j (\mu_j - p_{c,j})^2$
        \ENDFOR
        \STATE \hspace{0.5cm} $\hat y \leftarrow \mathrm{Softmax}(-d_0, -d_1)$
        \STATE \textbf{CPAC loss:}
        \STATE \hspace{0.5cm} $\mathcal{L}_{\mathrm{clf}} \leftarrow \mathrm{BCE}(y, \hat y)$
        \STATE \hspace{0.5cm} $\mathcal{L}_{\mathrm{scale}} \leftarrow \lambda_{\mathrm{scale}} \|\alpha\|^2$
        \STATE \hspace{0.5cm} $\mathcal{L}_{\mathrm{anchor}} \leftarrow \lambda_{\mathrm{anchor}}(\|\mathbf{p}_0 - \bar{\mu}_0\|^2 + \|\mathbf{p}_1 - \bar{\mu}_1\|^2)$
        \STATE \hspace{0.5cm} $\mathcal{L}_{\mathrm{CPAC}} \leftarrow \mathcal{L}_{\mathrm{clf}} + \mathcal{L}_{\mathrm{scale}} + \mathcal{L}_{\mathrm{anchor}}$
        \STATE Update $\Theta_{\mathrm{CPAC}}, \phi \leftarrow \Theta_{\mathrm{CPAC}}, \phi - \eta \nabla \mathcal{L}_{\mathrm{CPAC}}$
    \ENDFOR
    \STATE Evaluate Precision and Recall on $\mathcal{D}_{\mathrm{val}}$
    \STATE Compute validation score $S = 0.5\,\mathrm{Precision} + 0.5\,\mathrm{Recall}$
    \IF{$S > S_{\mathrm{best}}$}
        \STATE Save model checkpoint
        \STATE $S_{\mathrm{best}} \leftarrow S$
        \STATE Reset early-stopping counter
    \ELSE
        \STATE Increment early-stopping counter
    \ENDIF
    \IF{early-stopping counter $\ge p$}
        \STATE \textbf{break}
    \ENDIF
\ENDFOR
\end{algorithmic}
\color{black}
\end{algorithm}

\section{Experimental Results}
\label{sec:experiments}
Given the proposed methodologies, we delve into all the experiments with the models and methods listed in Section~\ref{sec:methodologies}. First we test the CPAC against the other four classifiers and the results obtained by other works. In all experiments in this section, we use $k=5$ neighbors for KNN, as this configuration consistently yielded the best performance across all tests. We then analyze and compare the results obtained with the VAE-GAN equipped with a classification head against other oversampling techniques. All reported metrics are computed on the held-out test set.

\subsection{Preliminary Results}
\begin{figure}[ht]
  \centering
  \begin{subfigure}[b]{0.48\textwidth}
    \centering
    \includegraphics[width=0.8\linewidth]{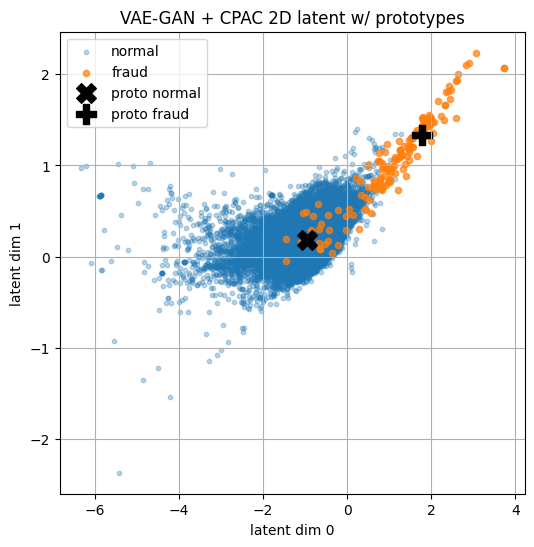}
    \caption{2D PCA: Encoder with CPAC head.}
    \label{fig:cpac_cluster}
  \end{subfigure}
  \hfill
  \begin{subfigure}[b]{0.48\textwidth}
    \centering
    \includegraphics[width=0.8\linewidth]{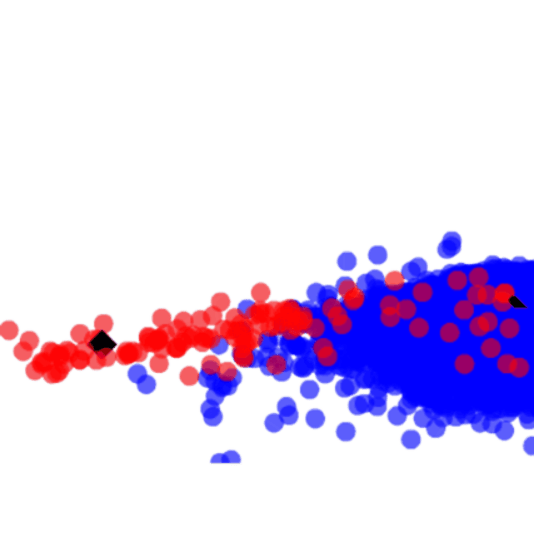}
    \caption{3D cluster overlap.}
    \label{fig:overlap1}
  \end{subfigure}
  \caption{Latent space visualizations for the Encoder with CPAC head: (a) PCA, (b) 3D overlap.}
  \label{fig:cpac_head_subfigs}
\end{figure}

Unlike MLP heads, the CPAC’s architecture and losses (especially prototype anchoring) push the encoder to organize latent means $\mu$ into two tight, well-separated clusters, each surrounding its prototype. The attention weights $\mathbf{w}$ add interpretability, highlighting which latent features are most influential for distinguishing frauds. This process results in a latent space that is both discriminative and interpretable. As we can see on Figure~\ref{fig:cpac_cluster} the CPAC was able to improve separation creating two distinct clusters with each prototype being anchored to its cluster centroid, essentially acting like one. A limited number of misclassifications is observed, which is expected in highly imbalanced scenarios, together with a small residual overlap between the two classes (visible in Figure~\ref{fig:overlap1}) which is caused by borderline transactions. This is the first step to a more aware encoder that knows what actually constitutes a fraud.

\subsection{Oversampling results and experiments}
Now, we will cover the results obtained with the two state-of-the-art oversampling strategies~\citep{chawla2002smote, kingma2014autoencoding} applied to Logistic Regression, Random Forest, XGBoost and CPAC. First they will be tested without oversampling, to asses how oversampling can  aid overall performances; then they will be tested with SMOTE oversampling and VAE-GAN.

\subsubsection{No Oversampling}
All models are first evaluated on the original, non-augmented dataset to establish a baseline, with results reported in Table~\ref{tab:baseline}. Despite the strong class imbalance, all methods achieve solid performance, indicating that the dataset already provides informative signals for fraud detection.
Logistic Regression shows balanced Precision and Recall but is slightly outperformed by more expressive models, reflecting its limited ability to capture complex decision boundaries. Random Forest attains the highest Precision at the expense of Recall, suggesting a conservative classification behavior. XGBoost achieves a strong balance between Precision and Recall, resulting in high F1-score and AUC-ROC values, and confirming its robustness on the original data distribution. KNN also performs competitively, though with a lower AUC-ROC, indicating weaker global ranking performance.
The CPAC classifier demonstrates well-balanced results, achieving high Recall and the highest AUC-ROC among all models. This suggests that its prototype-based structure effectively captures meaningful latent representations even without oversampling. While CPAC performs comparably to tree-based ensembles in this setting, it is likely to further benefit from enriched training data, where additional minority samples could help stabilize prototypes and refine decision boundaries.

\begin{table*}[!ht]
  \caption{Benchmark test results on the original (non-augmented) dataset. Results in bold indicate the best values, underlined ones represent the second best.}
  \label{tab:baseline}
  \centering
  \tiny
  \begin{tabular}{l r r r r}
    \toprule
    \textbf{Model}               & \textbf{Precision (\%)} & \textbf{Recall (\%)} & \textbf{F1-score (\%)} & \textbf{AUC-ROC (\%)} \\
    \midrule
    Logistic Regression & 86.77 & 89.77 & 88.21 & 97.35 \\
    Random Forest       & \textbf{96.65} & 85.25 & 90.59 & 97.87 \\
    XGBoost             & \underline{95.44} & \textbf{90.81} & \textbf{93.00} & \underline{98.14} \\
    KNN                 & 95.33 & \underline{89.79} & \underline{92.38} & 94.88 \\
    CPAC                & 93.16 & 88.76 & 90.85 & \textbf{98.49} \\
    \bottomrule
  \end{tabular}
\end{table*}

\color{black}
\subsubsection{Single-Sample Inference Time}
Inference efficiency is a critical requirement in real-world credit card fraud detection systems, where models are often deployed in latency-sensitive environments. Table~\ref{tab:inference_time} reports the average single-sample inference time measured on the test set for the different classifiers considered in this study, all trained without oversampling.
As expected, linear and tree-based models such as Logistic Regression and XGBoost exhibit extremely low inference latency, making them well suited for high-throughput real-time scoring. More instance-based methods such as KNN incur higher inference costs due to their reliance on neighborhood search at prediction time. The proposed CPAC model introduces a moderate computational overhead compared to traditional classifiers, while remaining significantly faster than KNN and well within practical latency constraints.
Importantly, CPAC is not intended to replace lightweight production scorers in isolation. Its primary role is to act as a \emph{classifier-guided regularizer during training}, shaping the latent space of the generative model through prototype anchoring and attention-weighted feature interactions. This latent structuring improves class separability, robustness, and downstream detection performance under extreme class imbalance. In practical deployment scenarios, CPAC can therefore be used either offline or within hybrid pipelines, while final real-time decisions may still rely on simpler classifiers operating on the improved representations.
Overall, these results indicate that the proposed approach achieves a favorable trade-off between representational benefits and computational cost, supporting its applicability in realistic fraud detection settings.

\begin{table*}[!htp]
\color{black}
\centering
\caption{Single-sample inference time on the test set for different classifiers trained without oversampling.}
\label{tab:inference_time}
\tiny
\begin{tabular}{l c}
\hline
\textbf{Classifier} & \textbf{Inference Time (ms)} \\
\hline
Logistic Regression & \textbf{0.000046} \\
Random Forest       & 0.003780 \\
KNN                 & 0.027479 \\
XGBoost             & \underline{0.000158} \\
CPAC                & 0.009852 \\
\hline
\end{tabular}
\color{black}
\end{table*}
\color{black}

\subsubsection{SMOTE Oversampling Performances}
Applying SMOTE oversampling with 50, 75, and 100 synthetic fraud samples leads to consistent performance changes across all classifiers, as reported in Table~\ref{tab:smote-benchmarks}. While Random Forest achieves very strong Precision and F1-score, particularly at higher oversampling levels, its performance varies noticeably across different augmentation sizes. This behavior suggests that the model is highly sensitive to the specific structure induced by the synthetic samples during training, resulting in decision boundaries that do not generalize consistently to the real, unseen test data.
In contrast, XGBoost demonstrates the most reliable behavior under SMOTE augmentation, maintaining a stable balance between Precision and Recall and achieving consistently high F1-score and AUC-ROC across all oversampling levels. This stability highlights XGBoost’s robustness to class imbalance and its ability to exploit synthetic augmentation without excessively distorting the learned decision function.
Logistic Regression benefits modestly from oversampling, with gradual improvements in Recall and F1-score, but remains constrained by its linear decision boundaries. CPAC shows competitive performance at lower oversampling levels, particularly in terms of AUC-ROC, but its metrics degrade as the number of synthetic samples increases, indicating sensitivity to changes in the training data distribution.
Overall, these results indicate that while SMOTE oversampling can enhance performance, especially for ensemble methods.

\begin{table*}[!htp]
  \caption{Benchmark test results using SMOTE oversampling with 50, 75, and 100 synthetic fraud samples. Results in bold indicate the best values, underlined ones represent the second best.}
  \label{tab:smote-benchmarks}
  \centering
  \tiny
  \begin{tabular}{r l r r r r}
    \toprule
    \textbf{\# Samples} & \textbf{Model} & \textbf{Precision (\%)} & \textbf{Recall (\%)} & \textbf{F1-score (\%)} & \textbf{AUC-ROC (\%)} \\
    \midrule
    50 & Logistic Regression & 86.77 & \underline{89.77} & 88.21 & 97.37 \\
    50 & Random Forest       & \textbf{96.65} & 85.23 & 90.58 & 97.87 \\
    50 & XGBoost             & 95.33 & \textbf{89.79} & \textbf{92.38} & \textbf{98.46} \\
    50 & KNN                 & 95.33 & \textbf{89.79} & \textbf{92.38} & 94.88 \\
    50 & CPAC                & \underline{96.32} & 88.77 & \underline{92.21} & \underline{98.13} \\
    \midrule
    75 & Logistic Regression & 86.35 & \underline{90.79} & 88.44 & 97.41 \\
    75 & Random Forest       & \textbf{96.78} & 86.51 & 91.35 & \underline{97.85} \\
    75 & XGBoost             & 94.55 & \textbf{91.83} & \textbf{93.15} & \textbf{98.25} \\
    75 & KNN                 & \underline{95.12} & 89.79 & \underline{92.38} & 94.88 \\
    75 & CPAC                & \underline{95.12} & 78.57 & 84.98 & 97.11 \\
    \midrule
    100 & Logistic Regression & 87.26 & 91.81 & 89.40 & 97.49 \\
    100 & Random Forest       & \textbf{97.66} & \textbf{91.83} & \textbf{94.56} & \underline{97.85} \\
    100 & XGBoost             & 94.43 & 90.81 & \underline{92.54} & \textbf{98.22} \\
    100 & KNN                 & \underline{95.33} & 89.79 & 92.38 & 94.88 \\
    100 & CPAC                & 90.99 & \underline{91.82} & 91.40 & 96.71 \\
    \bottomrule
  \end{tabular}
\end{table*}

\subsubsection{VAE--GAN Oversampling Performances}
VAE-GAN-based oversampling with 50, 75, and 100 synthetic fraud samples leads to consistent performance improvements across most classifiers, as reported in Table~\ref{tab:vaegan-benchmark}. Compared to SMOTE, the VAE-GAN generates higher-quality synthetic samples that more effectively support discriminative learning, particularly for ensemble models.
Random Forest achieves very strong Precision and F1-score across all oversampling levels; however, its performance varies with the number of synthetic samples, indicating a higher sensitivity to training-time distribution changes. In contrast, XGBoost exhibits the most reliable behavior, maintaining a stable balance between Precision and Recall and achieving consistently high F1-score and AUC-ROC across all configurations. This stability confirms XGBoost’s ability to exploit generative oversampling while preserving generalization to real test data.
KNN shows largely unchanged performance across different oversampling levels, mirroring the behavior observed with SMOTE. As an instance-based method, KNN remains primarily influenced by local neighborhood structure, and the additional synthetic samples, while plausible,do not substantially alter the nearest-neighbor geometry governing its predictions.
CPAC demonstrates competitive performance, particularly in terms of AUC-ROC, but its Recall and F1-score degrade as the number of synthetic samples increases, suggesting sensitivity to shifts in the latent data distribution. Logistic Regression continues to benefit only marginally from oversampling due to its limited representational capacity.
Overall, these results indicate that VAE-GAN oversampling provides more informative synthetic data than SMOTE, but its benefits are best realized by models with strong regularization and margin-aware learning.

\begin{table*}[!htp]
  \caption{Benchmark test results using a VAE-GAN oversampler with 50, 75, and 100 synthetic fraud samples. Results in bold indicate the best values, underlined ones represent the second best.}
  \label{tab:vaegan-benchmark}
  \centering
  \tiny
  \begin{tabular}{r l r r r r}
    \toprule
    \textbf{\# Samples} & \textbf{Model} & \textbf{Precision (\%)} & \textbf{Recall (\%)} & \textbf{F1-score (\%)} & \textbf{AUC-ROC (\%)} \\
    \midrule
    50 & Logistic Regression & 86.52 & 88.75 & 87.60 & 97.40 \\
    50 & Random Forest       & \textbf{97.60} & \textbf{90.81} & \textbf{93.95} & \underline{97.85} \\
    50 & XGBoost             & \underline{96.70} & 88.51 & 92.21 & 96.72 \\
    50 & KNN                 & 95.33 & \underline{89.79} & 92.38 & 94.88 \\
    50 & CPAC                & 96.32 & \underline{88.77} & 92.21 & 97.71 \\
    \midrule
    75 & Logistic Regression & 86.25 & 87.73 & 86.98 & 97.20 \\
    75 & Random Forest       & \textbf{97.66} & \textbf{91.83} & \textbf{94.56} & \underline{97.88} \\
    75 & XGBoost             & 94.55 & \textbf{91.83} & \underline{93.15} & \textbf{98.32} \\
    75 & KNN                 & 95.33 & \underline{89.79} & 92.38 & 94.88 \\
    75 & CPAC                & \underline{96.03} & 85.71 & 90.21 & 97.40 \\
    \midrule
    100 & Logistic Regression & 86.52 & 88.75 & 87.60 & 97.29 \\
    100 & Random Forest       & \textbf{96.58} & \textbf{91.83} & \textbf{94.08} & 96.84 \\
    100 & XGBoost             & 94.67 & \underline{91.61} & \underline{93.11} & \underline{98.12} \\
    100 & KNN                 & \underline{95.33} & 89.79 & 92.38 & 94.88 \\
    100 & CPAC                & 89.56 & 88.76 & 89.16 & \textbf{98.43} \\
    \bottomrule
  \end{tabular}
\end{table*}

\subsubsection{SMOTE vs VAE-GAN Results Observations}
Comparative analysis shows that VAE–GAN-generated synthetic frauds provide greater improvements in recall and F1-score than SMOTE, particularly for XGBoost and CPAC. SMOTE easily fills gaps within clusters but can generate less realistic points at class boundaries. VAE–GAN, while more complex, generates realistic samples that support better model generalization. When paired with VAE–GAN, CPAC shows pronounced gains in recall and AUC-ROC, especially with more synthetic samples. However, excessive oversampling can lead to plateauing or overfitting, highlighting the need to find an optimal level of augmentation.

\subsection{VAE-GAN with CPAC Head as Oversampler}
Table~\ref{tab:vaegancpac-benchmark} shows the results obtained using the VAE-GAN+CPAC to generate synthetic fraud samples. Compared to previous oversampling strategies, performance gains are more moderate and quickly plateau as the number of synthetic samples increases, suggesting that the generated data closely matches the real fraud distribution and avoids aggressive overfitting.
Logistic Regression benefits only marginally from this oversampling strategy, while Random Forest achieves strong Precision but exhibits limited performance variation across oversampling levels. XGBoost demonstrates the most reliable behavior, maintaining a stable balance between Precision and Recall and achieving consistently high F1-score and AUC-ROC across all configurations. KNN remains largely unaffected due to its instance-based nature. Overall, the VAE-GAN+CPAC oversampler favors stable and generalizable learning, with XGBoost emerging as the most suitable downstream classifier.

\begin{table}[!htp]
  \caption{Benchmark test results using a VAE-GAN+CPAC oversampler with 50, 75, and 100 synthetic fraud samples. Results in bold indicate the best values, underlined ones represent the second best.}
  \label{tab:vaegancpac-benchmark}
  \centering
  \tiny
  \begin{tabular}{r l r r r r}
    \toprule
   \textbf{\# Samples} & \textbf{Model} & \textbf{Precision (\%)} & \textbf{Recall (\%)} & \textbf{F1-score (\%)} & \textbf{AUC-ROC (\%) }\\
    \midrule
    50 & Logistic Regression & 86.25 & 89.77 & 87.48 & \underline{97.05} \\
    50 & Random Forest       & \textbf{96.58} & \textbf{91.83} & \textbf{94.08} & 96.85 \\
    50 & XGBoost             & 94.55 & \textbf{91.83} & \underline{93.15} & \textbf{98.49} \\
    50 & KNN                 & \underline{95.33} & \underline{89.79} & 92.38 & 94.88 \\
    \midrule
    75 & Logistic Regression & 85.83 & 88.75 & 87.23 & \underline{97.59} \\
    75 & Random Forest       & \textbf{96.58} & \textbf{91.83} & \textbf{94.08} & 96.85 \\
    75 & XGBoost             & 94.43 & \underline{90.81} & \underline{92.54} & \textbf{98.31} \\
    75 & KNN                 & \underline{95.33} & 89.79 & 92.38 & 94.88 \\
    \midrule
    100 & Logistic Regression & 86.09 & 89.77 & 87.84 & \underline{97.38} \\
    100 & Random Forest       & \textbf{96.50} & \underline{90.81} & \textbf{93.47} & 96.81 \\
    100 & XGBoost             & 94.55 & \textbf{91.83} & \underline{93.15} & \textbf{98.89} \\
    100 & KNN                 & \underline{95.33} & 89.79 & 92.38 & 94.88 \\ 
    \bottomrule
  \end{tabular}
\end{table}

\subsubsection{Applying Oversampling before VAE-GAN+CPAC Training}
\begin{figure}[!ht]
  \centering
  \begin{subfigure}[b]{0.48\linewidth}
    \centering
    \includegraphics[width=\linewidth]{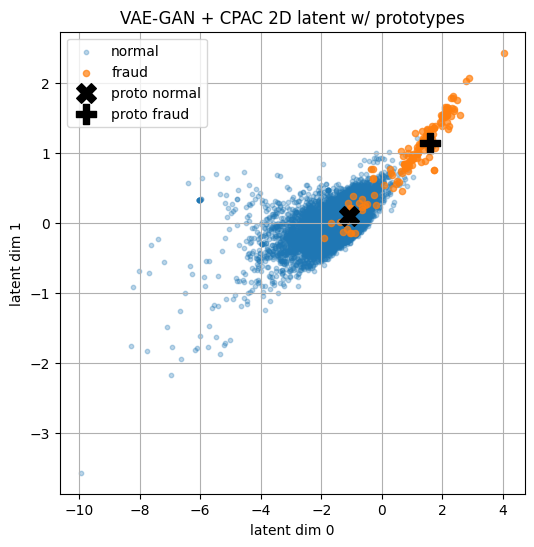}
    \caption{2D PCA: CPAC head, SMOTE pre-oversampling (75 samples).}
    \label{fig:cpac_cluster_ovs}
  \end{subfigure}
  \hfill
  \begin{subfigure}[b]{0.48\linewidth}
    \centering
    \includegraphics[width=\linewidth]{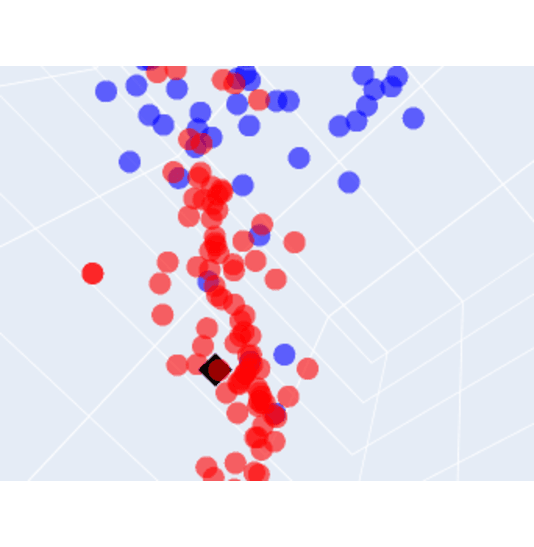}
    \caption{3D cluster overlap: SMOTE pre-oversampling (75 samples).}
    \label{fig:overlap2}
  \end{subfigure}
  \caption{Latent space visualizations with CPAC head and SMOTE pre-oversampling. (a) PCA, (b) 3D overlap.}
  \label{fig:cpac_ovs_subfigs}
\end{figure}

\begin{figure}[!ht]
  \centering
  \includegraphics[width=0.5\linewidth]{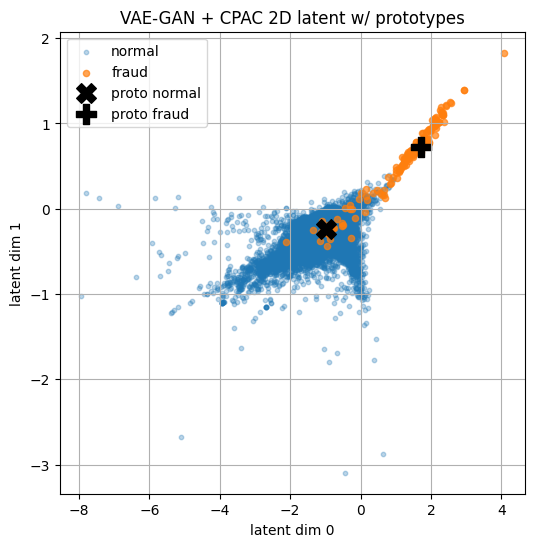}
  \caption{2D PCA plot of the cluster representation of the Encoder with CPAC head trained with SMOTE pre-oversampling over 100 samples.}
  \label{fig:cpac_cluster_ovs100}
\end{figure}

To further improve latent space separation in the VAE-GAN+CPAC framework, we explored the effect of applying a slight oversampling to the fraud class before pipeline training. Specifically, we employed SMOTE, as it produces a broader spread of synthetic fraud samples across the cluster, offering a better foundation for the encoder to learn generalizable boundaries. Although this choice may appear counter-intuitive in light of our previous discussion on minority-only oversampling, in this case we need a slight "accentuation" of the frauds samples defined in the overlap in order to allow the model to detect them better and  further reduce the overlap. 
Pre-training augmentation with VAE-GAN tended to generate samples in a narrower region of the minority class, providing less diversity and thus limited improvement in latent separation. Guided by previous experiments, we selected 75 SMOTE-generated samples as the optimal amount, since further augmentation did not yield additional gains (see Figure~\ref{fig:cpac_cluster_ovs100}). This approach visibly reduced cluster overlap in the latent space (Figures~\ref{fig:cpac_cluster_ovs},~\ref{fig:overlap2}) and improved overall model performance. Thus, SMOTE-based pre-training oversampling proved more effective than VAE-GAN augmentation in this context, with 75 synthetic samples offering the best balance between diversity and separation.

\subsubsection{Results Obtained with Pre-Training Oversampled VAE-GAN+CPAC}
Table~\ref{tab:vaegancpacovs-benchmark} reports the results obtained when the VAE-GAN+CPAC pipeline is trained on a pre-oversampled dataset. Compared to the standard VAE-GAN+CPAC configuration, this strategy yields clearer and more consistent performance gains, particularly for XGBoost, which achieves the highest F1-score and AUC-ROC across all oversampling levels.
Random Forest continues to obtain very strong Precision and competitive F1-scores, but its performance varies with the number of synthetic samples. In contrast, XGBoost demonstrates superior stability, maintaining a well-balanced Precision-Recall trade-off and consistently high AUC-ROC, indicating improved generalization to real test data. Logistic Regression benefits only marginally, while KNN remains largely unaffected.
Overall, pre-training oversampling further enhances the effectiveness of the VAE-GAN+CPAC pipeline by producing cleaner latent representations that are best exploited by robust, margin-aware classifiers, with XGBoost emerging as the most reliable downstream model.

\begin{table}[!htp]
  \caption{Benchmark test results using a pre-training oversampled VAE-GAN+CPAC oversampler with 50, 75, and 100 synthetic fraud samples. Results in bold indicate the best values, underlined ones represent the second best.}
  \label{tab:vaegancpacovs-benchmark}
  \centering
  \tiny
  \begin{tabular}{r l r r r r}
    \toprule
    \textbf{\# Samples} & \textbf{Model} & \textbf{Precision (\%)} & \textbf{Recall (\%)} & \textbf{F1-score (\%)} & \textbf{AUC-ROC (\%)} \\
    \midrule
    50 & Logistic Regression & 86.25 & 87.73 & 86.98 & 97.36 \\
    50 & Random Forest       & \textbf{96.65} & \textbf{92.85} & \textbf{94.67} & \underline{97.86} \\
    50 & XGBoost             & 92.84 & \underline{92.84} & \underline{92.94} & \textbf{98.66} \\
    50 & KNN                 & \underline{95.33} & 89.79 & 92.38 & 94.88 \\
    \midrule
    75 & Logistic Regression & 84.80 & 89.77 & 87.12 & \underline{97.55} \\
    75 & Random Forest       & \textbf{97.60} & \underline{90.81} & \textbf{93.95} & 96.84 \\
    75 & XGBoost             & 93.74 & \textbf{92.85} & \underline{93.29} & \textbf{98.77} \\
    75 & KNN                 & \underline{95.33} & 89.79 & 92.38 & 94.88 \\
    \midrule
    100 & Logistic Regression & 85.07 & 90.79 & 87.71 & 97.42 \\
    100 & Random Forest       & \textbf{96.50} & \underline{90.81} & \underline{93.47} & \underline{97.87} \\
    100 & XGBoost             & 94.67 & \textbf{92.85} & \textbf{93.74} & \textbf{98.47} \\
    100 & KNN                 & \underline{95.33} & 89.79 & 92.38 & 94.88 \\
    \bottomrule
  \end{tabular}
\end{table}

\subsection{Comparison with State of the Art Approaches}
To evaluate the effectiveness of our approach, we benchmarked the best model obtained in this work (XGBoost) trained with pre-training oversampled VAE-GAN+CPAC data against recent state-of-the-art methods~\citep{Ding2023, Shi2025_balVAE, Ahmed2025}. We conducted these experiments on these works instead of others listed in Section~\ref{sec:related_work} because they are more recent. Since the code for the selected works is not available we reproduced their classification settings at the best of our abilities and accordingly to the information given; Ding et al.~\citep{Ding2023} uses an XGBoost as a baseline classifier just like us, so the reported results of our models reflects theirs with our generated dataset proving that our method greatly improves the performances. Shi et al.~\citep{Shi2025_balVAE} uses a multi-head attention classifier in their generative pipeline to directly classify the samples reporting perfect metrics, so, we took just the classifier to test with our generated data. Ahmed et al.~\citep{Ahmed2025} uses a voting ensemble classification method that comprises a Random Forest Classifier, AdaBoost Classifier and KNN, similary reporting perfect metrics. The Table~\ref{tab:oursvtheirs} reports the evaluation results of each method compared to ours with their relative setup: Shi et al. uses a 75/25 split for the dataset and balances the training set so that the minority class reaches the same number of the majority class. Ahmed et al. uses a 80/20 split and balances the training set as well. We ran both tests and compared the results with each method and our XGBoost outperforms in every metric. Table~\ref{tab:oursvtheirs1} reports the results obtained with our setup (80/10/10 split and only 100 generated samples in the training set) and even in this instance our model outperforms the other proposed methodologies. The main difference between our proposed method of training and the related works's is that we do not balance the training set because we believe that the models can only benefit from learning an extreme imbalanced distribution during training, adapting it to real world applications even appropriately.


\begin{table}[!htp]
    \centering
    \caption{Benchmark XGBoost evaluation performances trained with pre-training oversampled VAE-GAN+CPAC data, compared to selected recent works with their setup and splits.}
    \tiny
    \begin{tabular}{@{}lcccc@{}}
    \toprule
    \textbf{Work} & \textbf{Precision (\%)} & \textbf{Recall (\%)} & \textbf{F1-score (\%)} & \textbf{AUC-ROC (\%)} \\
    \midrule
    Shi et al.~\cite{Shi2025_balVAE} (2025)   & 79.50  & 78.86  & 79.18  & 97.56  \\
    Ours                                      & \textbf{97.58}  & \textbf{90.24}  & \textbf{93.60}  & \textbf{97.83}  \\
    \midrule
    Ahmed et al.~\cite{Ahmed2025} (2025)      & 93.75  & 76.53  & 84.27  & 97.37  \\
    Ours                                      & \textbf{95.44}  & \textbf{90.81}  & \textbf{93.00}  & \textbf{98.10}  \\
    \bottomrule
    \end{tabular}
    \label{tab:oursvtheirs}
\end{table}

\begin{table}[!htp]
    \centering
    \caption{Benchmark XGBoost evaluation performances trained with pre-training oversampled VAE-GAN+CPAC data at 100 samples, compared to selected recent works with our setup.}
    \tiny
    \begin{tabular}{@{}lcccc@{}}
    \toprule
    \textbf{Work} & \textbf{Precision (\%)} & \textbf{Recall (\%)} & \textbf{F1-score (\%)} & \textbf{AUC-ROC (\%)} \\
    \midrule
    Shi et al.~\cite{Shi2025_balVAE} (2025)   & 78.76  & 77.70  & 78.23  & 97.56  \\
    Ours                                      & \textbf{94.67}  & \textbf{92.85}  & \textbf{93.74}  & \textbf{98.47}  \\
    \midrule
    Ahmed et al.~\cite{Ahmed2025} (2025)      & 94.59  & 70.95  & 81.08  & 96.29  \\
    Ours                                      & \textbf{94.67}  & \textbf{92.85}  & \textbf{93.74}  & \textbf{98.47}  \\
    \bottomrule
    \end{tabular}
    \label{tab:oursvtheirs1}
\end{table}

\subsection{Evaluation on an Independent Synthetic Credit Card Dataset}
To further assess the generalization capability of the proposed VAE-GAN+CPAC oversampler, we evaluate our approach on an additional credit card fraud detection dataset publicly available on Kaggle and published by Kartik Shenoy~\citep{shenoy2020dataset}. The use of an independent dataset is particularly important in fraud detection, where the availability of publicly accessible and well-documented datasets is extremely limited due to privacy and regulatory constraints.
The dataset is fully synthetic and was generated through a controlled simulation process, providing a viable alternative for cross-dataset validation. It consists of simulated credit card transactions spanning the period from January 1, 2019 to December 31, 2020, modeling the activity of 1,000 customers interacting with a pool of 800 merchants. Both legitimate and fraudulent transactions are included. The data were generated using the Sparkov Data Generation tool~\citep{harris2022tool}, an open-source simulator designed to emulate realistic credit card usage patterns over time. The individual simulation outputs were subsequently aggregated and converted into a standardized format.
Importantly, the dataset is distributed with a predefined and fixed train-test split, which we preserve throughout all experiments. This ensures a fair and reproducible evaluation, prevents any form of data leakage, and allows us to isolate the effect of oversampling techniques, as synthetic data generation and model training are strictly confined to the training partition, while all reported results are computed exclusively on the untouched test set.

\subsubsection{Subsequent Results}
\begin{figure}[!ht]
  \centering
  \includegraphics[width=0.5\linewidth]{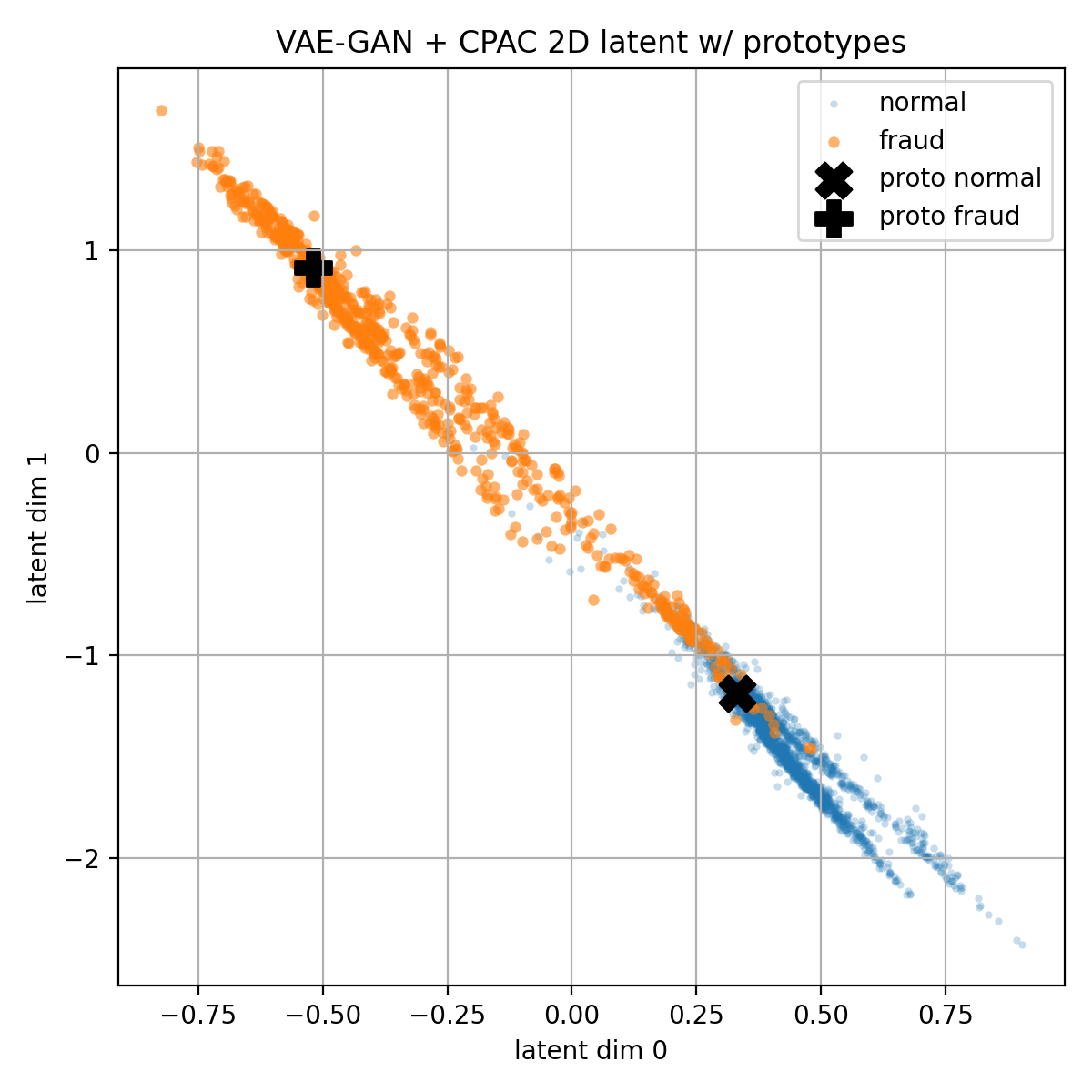}
  \caption{Latent space visualizations for the Encoder with CPAC head obtained with the Synthetic Dataset. The two clusters show little overlap and a clean separation.}
  \label{fig:vaegan_new_dataset}
\end{figure}

To evaluate the proposed VAE-GAN+CPAC framework on the newly introduced dataset, we first trained the model without any pre-training oversampling to assess its behavior on clean, untouched data. Even under this setting, the encoder learns a well-structured latent representation, as illustrated in Figure~\ref{fig:vaegan_new_dataset}, where fraud and non-fraud samples form clearly separated clusters with minimal overlap, a characteristic expected in highly imbalanced scenarios. The learned prototypes are positioned far apart in the latent space, further confirming the model’s ability to capture discriminative structure without relying on synthetic augmentation.
Building on this baseline, we progressively augmented the training set with 50, 75, and 100 synthetic fraud samples generated by the VAE-GAN+CPAC oversampler. The downstream classifiers were then trained on the augmented data and evaluated exclusively on the predefined test set, which remains untouched by any synthetic samples. We first report the results of the classifiers without any oversampling in Table~\ref{tab:baseline_new_dataset} as a reference.
The quantitative results are reported in Table~\ref{tab:vaegancpac-newdataset}. Ensemble methods consistently outperform simpler classifiers across all oversampling levels. Random Forest achieves strong Precision and competitive F1-scores, but its performance fluctuates with the number of synthetic samples. In contrast, XGBoost exhibits the most reliable behavior, achieving the highest F1-score and AUC-ROC at higher augmentation levels while maintaining a stable Precision-Recall balance across all configurations.
Logistic Regression and KNN show limited sensitivity to the oversampling process, reflecting their constrained representational capacity and instance-based nature, respectively. Overall, these results confirm that the VAE-GAN+CPAC oversampler generalizes effectively to a distinct synthetic dataset and that its benefits are most consistently exploited by robust, margin-aware classifiers such as XGBoost.

\begin{table*}[!ht]
  \caption{Benchmark test results on the new (non-augmented) dataset. Results in bold indicate the best values, underlined ones represent the second best.}
  \label{tab:baseline_new_dataset}
  \centering
  \tiny
  \begin{tabular}{l r r r r}
    \toprule
    \textbf{Model}               & \textbf{Precision (\%)} & \textbf{Recall (\%)} & \textbf{F1-score (\%)} & \textbf{AUC-ROC (\%)} \\
    \midrule
    Logistic Regression & 65.06 & 66.96 & 65.95 & 82.79 \\
    Random Forest       & \textbf{95.47} & \textbf{88.75} & \textbf{91.84} & \underline{97.71} \\
    XGBoost             & \underline{89.93} & \underline{86.07} & \underline{87.86} & \textbf{99.63} \\
    KNN                 & 75.95 & 77.20 & 76.56 & 81.98 \\
    \bottomrule
  \end{tabular}
\end{table*}

\begin{table}[!htp]
  \caption{Benchmark test results using a VAE-GAN+CPAC oversampler with 50, 75, and 100 synthetic fraud samples on the newly introduced dataset. Results in bold indicate the best values, underlined ones represent the second best.}
  \label{tab:vaegancpac-newdataset}
  \centering
  \tiny
  \begin{tabular}{r l r r r r}
    \toprule
   \textbf{\# Samples} & \textbf{Model} & \textbf{Precision (\%)} & \textbf{Recall (\%)} & \textbf{F1-score (\%)} & \textbf{AUC-ROC (\%) }\\
    \midrule
    50 & Logistic Regression & 65.11 & 66.47 & 65.76 & 82.60 \\
    50 & Random Forest       & \textbf{93.40} & \textbf{89.67} & \textbf{91.45} & \underline{97.33} \\
    50 & XGBoost             & \underline{90.79} & \underline{85.42} & \underline{87.92} & \textbf{99.65} \\
    50 & KNN                 & 75.89 & 77.20 & 76.53 & 81.98 \\
    \midrule
    75 & Logistic Regression & 65.07 & 67.12 & 66.03 & 82.78 \\
    75 & Random Forest       & \textbf{95.13} & \textbf{88.82} & \textbf{91.74} & \underline{95.89} \\
    75 & XGBoost             & \underline{90.28} & \underline{87.03} & \underline{88.59} & \textbf{99.63} \\
    75 & KNN                 & 75.89 & 77.20 & 76.53 & 81.98 \\
    \midrule
    100 & Logistic Regression & 65.14 & 66.68 & 65.87 & 82.64 \\
    100 & Random Forest       & \textbf{94.61} & \underline{89.07} & \underline{91.66} & \underline{95.00} \\
    100 & XGBoost             & \underline{93.74} & \textbf{90.13} & \textbf{91.90} & \textbf{99.63} \\
    100 & KNN                 & 75.89 & 77.20 & 76.53 & 81.98 \\
    \bottomrule
  \end{tabular}
\end{table}

\section{Ablation Study}
\label{sec:ablation}
In order to establish the relevance of each component in our method, we systematically analyze the contribution of each key component in our VAE-GAN+CPAC architecture. By selectively removing or disabling certain elements, we demonstrate that every part of the model is essential for achieving discriminative and robust latent representations, particularly in the context of extreme class imbalance.

\subsection{Effect of Removing the CPAC Head}
\begin{figure}[!ht]
  \centering
  \includegraphics[width=0.5\textwidth]{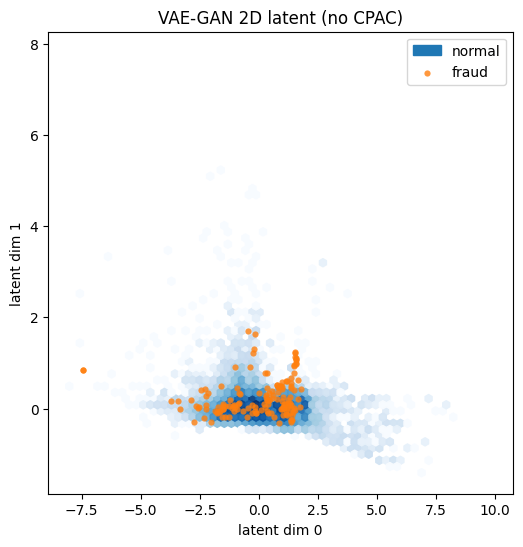}
  \caption{2D PCA plot of the cluster representation of the Encoder without CPAC. \color{black}Lower density points are less colored, while higher density (bigger overlap) are more highlighted.\color{black}}
  \label{fig:vaegan_clusters}
\end{figure}

To evaluate the impact of the CPAC head, we trained the VAE-GAN with the same settings as our main pipeline, but without the CPAC head. In this configuration, the encoder is updated only via generative objectives, with no explicit supervision guiding the latent space. Our results show that (as shown in Figure~\ref{fig:vaegan_clusters}), without the classifier, the latent representations of fraud and normal transactions remain heavily overlapped, making downstream classification significantly less accurate. This experiment confirms that the CPAC supervision is crucial for inducing clear separation between classes and for shaping the latent space in a way that supports reliable detection.

\subsection{Effect of Removing the Attention Mechanism from CPAC}
\begin{figure}[!ht]
  \centering
  \includegraphics[width=0.5\textwidth]{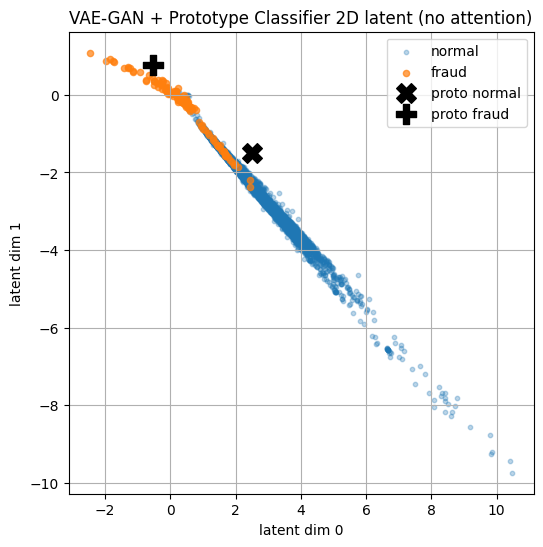}
  \caption{2D PCA plot of the cluster representation of the Encoder of VAEGAN+CPAC without the attention mechanism.}
  \label{fig:cpac_no_attention}
\end{figure}

The ablation in Figure~\ref{fig:cpac_no_attention} illustrates the effect of disabling the attention mechanism in the CPAC, reducing it to a pure prototype-based classifier. Without feature-wise attention, each latent dimension is weighted equally when computing distances to the class prototypes. The resulting latent space is visibly less expressive: data points for both normal and fraud classes collapse along a near-linear manifold, with limited separation between the classes and their prototypes.
This collapse indicates that the model is unable to adaptively emphasize the most discriminative latent features, leading to suboptimal cluster separation and reduced interpretability. Both class prototypes tend to be positioned close to the respective cluster means, but the lack of per-dimension weighting prevents effective partitioning of ambiguous or borderline samples. As a result, the discriminative power of the model is noticeably diminished compared to the full CPAC, where the attention mechanism enables more nuanced, non-linear separation of minority-class samples.

\subsection{Effect of Removing the Prototypes from CPAC}
\begin{figure}[!ht]
  \centering
  \includegraphics[width=0.5\textwidth]{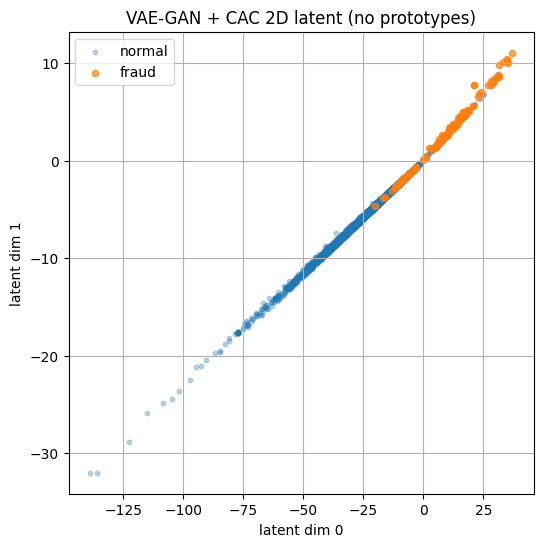}
  \caption{2D PCA plot of the cluster representation of the Encoder of VAEGAN+CPAC without prototypes.}
  \label{fig:cpac_no_prototypes}
\end{figure}

Figure~\ref{fig:cpac_no_prototypes} shows the latent space organization when the CPAC architecture is ablated to remove its prototype mechanism, leaving only the attention branch. In this setting, class discrimination relies exclusively on feature-wise weighting, with no explicit anchoring to learned class prototypes.
The resulting latent representations display a pronounced collapse along a narrow, near-linear manifold, with both normal and fraud samples largely overlapping. The absence of prototypes deprives the classifier of distinct, class-specific anchors in the latent space, severely restricting its ability to drive separation between classes. Although the attention branch allows the model to adaptively weight latent features, this alone proves insufficient for robust clustering, especially in the presence of extreme class imbalance.
Consequently, the discriminative structure of the latent space deteriorates, making class boundaries ambiguous and reducing interpretability. 

\subsection{Effect of Removing the Anchor and Scale Penalties}
\begin{figure}[!ht]
  \centering
  \includegraphics[width=0.5\textwidth]{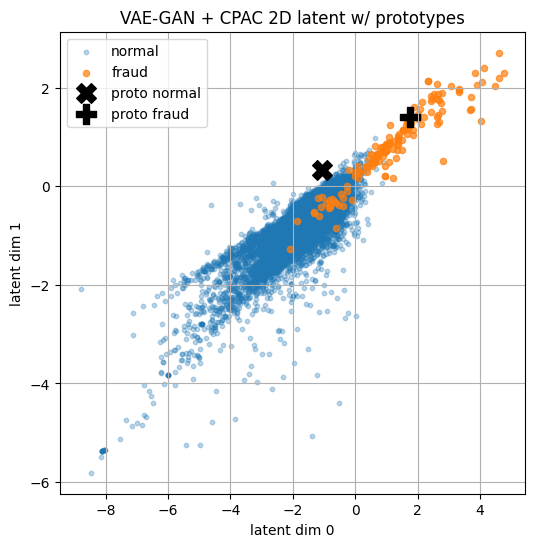}
  \caption{2D PCA plot of the cluster representation of the Encoder of VAEGAN+CPAC without scale and anchor penalties.}
  \label{fig:cpac_no_penalties}
\end{figure}

Figure~\ref{fig:cpac_no_penalties} visualizes the latent space structure when the CPAC classifier is trained without the scale and anchor regularization terms. In this setting, both the attention mechanism and learnable class prototypes remain active, but the model no longer receives explicit constraints on the spread and positioning of prototypes with respect to the cluster means.
The resulting latent representations maintain a moderate level of separation between normal and fraud clusters, with prototypes positioned near the centers of their respective class distributions. However, the boundaries between clusters are less crisp than in the fully regularized setting, and the spread of both clusters along the main latent direction increases. The absence of scale and anchor penalties allows prototypes to drift from the empirical cluster centers and can result in more diffuse class boundaries, reducing both interpretability and the sharpness of latent cluster assignments.

\subsection{Effect of Using Focal Loss for the CPAC}
\begin{figure}[!ht]
  \centering
  \begin{subfigure}[b]{0.48\textwidth}
    \centering
    \includegraphics[width=\textwidth]{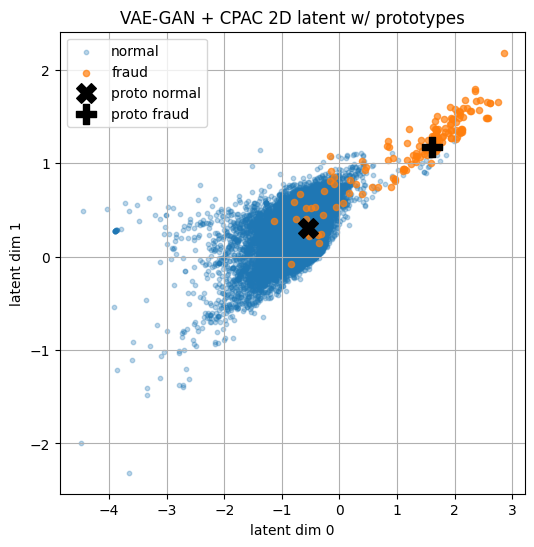}
    \caption{2D PCA plot of the cluster representation of the Encoder of VAEGAN+CPAC using Focal Loss for CPAC.}
    \label{fig:cpac_focal_latent}
  \end{subfigure}
  \hfill
  \begin{subfigure}[b]{0.48\textwidth}
    \centering
    \includegraphics[width=\textwidth]{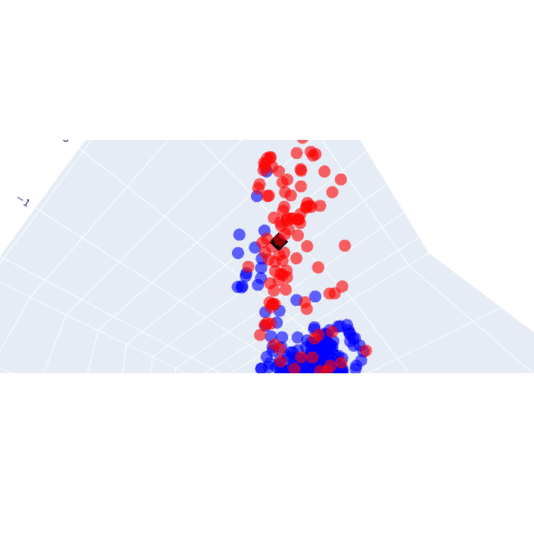}
    \caption{3D plot visualisation of the overlap for the Encoder of VAEGAN+CPAC using Focal Loss for CPAC.}
    \label{fig:cpac_focal_3d}
  \end{subfigure}
  \caption{Latent space visualizations for VAE-GAN+CPAC using Focal Loss: (a) 2D PCA plot; (b) 3D overlap visualization.}
  \label{fig:cpac_focal_subfigs}
\end{figure}

To further explore these ablations, we trained the VAE-GAN+CPAC pipeline using Focal Loss in place of standard binary cross-entropy as a loss function for the CPAC. As shown in Figure~\ref{fig:cpac_focal_latent}, this modification results in a more distinct and well-separated clustering of fraud (minority) and normal (majority) transactions in the latent space, with prototypes more cleanly anchoring their respective clusters. The 3D latent visualization (Figure~\ref{fig:cpac_focal_3d}) further confirms improved class-wise separation and greater dispersion of the minority class, compared to previous experiments using BCE (see Section~\ref{sec:experiments}).
Despite this evident structural improvement, Table~\ref{tab:vaegancpacfl-benchmark} reveals a slight decrease in downstream classification metrics across all evaluated classifiers and augmentation regimes. For instance, with 100 synthetic fraud samples, F1-scores for XGBoost and Random Forest remain above 92\%, but are marginally lower than those achieved with BCE-based training. This modest drop in quantitative performance can be attributed to the nature of Focal Loss: by prioritizing hard-to-classify minority samples, it encourages the model to push fraud examples away from the decision boundary, at the expense of global average accuracy and sometimes increased variability in the majority class predictions.

\begin{table}[!htp]
  \caption{Benchmark test results using VAE--GAN+CPAC oversampler with Focal Loss with 50, 75, and 100 synthetic fraud samples. Results in bold indicate the best values, underlined ones represent the second best.}
  \label{tab:vaegancpacfl-benchmark}
  \centering
  \tiny
  \begin{tabular}{r l r r r r}
    \toprule
    \textbf{\# Samples} & \textbf{Model} & \textbf{Precision (\%)} & \textbf{Recall (\%)} & \textbf{F1-score (\%)} & \textbf{AUC-ROC (\%)} \\
    \midrule
    50 & Logistic Regression & 84.80 & 89.77 & 87.12 & \underline{97.41} \\
    50 & Random Forest       & \textbf{96.65} & \underline{90.84} & \textbf{93.65} & 96.85 \\
    50 & XGBoost             & 93.60 & \textbf{91.83} & \underline{92.70} & \textbf{98.70} \\
    50 & KNN                 & \underline{95.33} & 89.79 & 92.38 & 94.88 \\
    \midrule
    75 & Logistic Regression & 86.09 & 89.77 & 87.84 & \underline{97.26} \\
    75 & Random Forest       & \textbf{96.58} & \textbf{91.83} & \textbf{94.08} & 96.85 \\
    75 & XGBoost             & 94.41 & \underline{89.99} & 92.15 & \textbf{98.42} \\
    75 & KNN                 & \underline{95.33} & 89.79 & \underline{92.38} & 94.88 \\
    \midrule
    100 & Logistic Regression & 89.51 & 84.68 & 86.94 & 97.56 \\
    100 & Random Forest       & \textbf{97.66} & \underline{90.01} & \underline{92.75} & \textbf{98.64} \\
    100 & XGBoost             & 94.30 & \textbf{91.51} & \textbf{92.88} & \underline{98.57} \\
    100 & KNN                 & \underline{95.33} & 89.79 & 92.38 & 94.88 \\
    \bottomrule
  \end{tabular}
\end{table}

\section{Discussion: Rethinking Oversampling in Fraud Detection}
\label{sec:discussion}
Despite significant advances in generative oversampling, most notably with SMOTE and VAE-GAN variants, most current state-of-the-art approaches in credit card fraud detection adopt a common strategy: training oversamplers exclusively on the minority (fraud) class. This “minority-only” paradigm is widespread, underpinned by the rationale that a focused model can better capture rare fraudulent patterns and help rebalance the dataset.
However, our results suggest that this strategy may have important limitations when applied to real-world fraud detection. Training an oversampler solely on fraud data risks generating synthetic samples that closely mimic observed frauds, rather than learning the nuanced differences between normal and fraudulent activity. As a result, such generated data may reflect interpolations within the minority class, lacking the true discriminative boundaries that are critical for effective classification. Deep and more complex classifiers rather than standard ones trained on these oversampled datasets often display overly optimistic metrics, sometimes failing to generalize robustly to new, unseen data or even overfitting even with few generated samples; simpler and classic classification approaches like the ones we used in this work struggle to improve their metrics especially in term of recall given the similarity between the synthetic frauds and the real ones.
Both SMOTE and vanilla VAE–GAN oversamplers succeed in enriching the minority class and boosting downstream classifier scores. However, SMOTE’s impressive precision and recall are largely artifacts of its linear interpolation, which generates highly similar fraud examples and trains classifiers to be overconfident within a narrow region of feature space.
The unsupervised VAE–GAN, while producing more realistic fraud instances, still suffers from overconfidence and mode collapse, generating synthetic cases that closely mimic the original frauds, limiting generalization.
When applying our CPAC classifier to these augmented datasets, we confirm a shared limitation: traditional oversamplers simply replicate, rather than expand, the fraud distribution.
In contrast, our proposed VAE-GAN+CPAC pipeline is designed to address these limitations. By training on the entire dataset, including both fraud and normal transactions, and integrating a CPAC head to provide explicit, supervised class information, our approach encourages the encoder to structure its latent space to distinguish fraud from non-fraud. This design is not simply an auxiliary feature, but a central objective of the model: the classifier head directly shapes the latent representations so that generated synthetic frauds meaningfully reflect the learned class boundaries. During inference, only the minority samples are generated, but the key insight is that the model’s holistic training allows it to create more realistic and discriminative synthetic data.
This approach yields several practical advantages, as reflected in our benchmarks:
\begin{itemize}
    \item \textbf{Reduced Overfitting and Overconfidence:} Models trained with our VAE-GAN+CPAC-generated frauds exhibit more stable and realistic performance, avoiding the inflated precision or recall sometimes observed with traditional oversamplers.
    \item \textbf{Improved Generalization:} The modest drop in certain metrics compared to some SOTA methods is, in fact, evidence of better generalization. Our synthetic frauds help downstream classifiers learn the true structure of the data, rather than simply memorizing training examples.
    \item \textbf{No Plateau Effect:} While standard oversamplers quickly reach a performance plateau or even degrade as more synthetic samples are added, our approach enables incremental improvements until larger sample sizes begin to induce overfitting, as expected.
    \item \textbf{Cluster Separation and Interpretability:} The explicit supervision provided by CPAC results in clearer separation between fraud and non-fraud clusters in latent space, facilitating interpretability and model transparency.
\end{itemize}

A key insight is that fraud can only be understood in the context of normal transactions, its definition is inherently relational. Oversamplers that ignore this context may generate synthetic data that resides within the convex hull of observed frauds, without sufficiently capturing the critical distinctions needed for robust classification. This can result in models that are prone to memorization, rather than effective discrimination.

\color{black}
\subsection{Operational Considerations and Business Impact}
In real-world fraud detection systems, performance must be interpreted not only in terms of aggregate metrics, but also with respect to operational costs and customer experience. In particular, false positives may lead to transaction friction, customer dissatisfaction, and increased manual review effort. The proposed CPAC-guided approach is designed to favor a recall-oriented behavior, prioritizing the identification of fraudulent transactions while maintaining a controlled increase in false positives.
This design choice reflects common industrial practice, where the cost of missed fraud typically outweighs the cost of additional verification steps. The use of an explicit decision threshold allows practitioners to tune the precision-recall trade-off according to domain-specific risk tolerance and business constraints. Compared to traditional oversampling strategies such as SMOTE, CPAC provides a more structured and stable decision boundary, resulting in improved recall without relying on aggressive oversampling that may amplify noise.
From a deployment perspective, CPAC primarily serves as a training-time mechanism to shape latent representations via prototype anchoring and attention, improving robustness under extreme class imbalance. In practical systems, this enables hybrid pipelines in which the improved representations can be paired with lightweight classifiers for real-time scoring, while preserving the interpretability and robustness benefits introduced during training.
\color{black}

\subsection{Our Contribution}
By training the VAE-GAN+CPAC on the full dataset and using the CPAC head to guide the latent representation, we bridge the gap between oversampling and supervised discriminative learning. The synthetic frauds produced by our approach are not naive copies, but instead reflect meaningful, learned differences between classes. This allows downstream classifiers, particularly XGBoost in our experiments, to achieve strong, generalizable performance, even under extreme class imbalance.
In summary, our findings support the view that the field would benefit from moving beyond the traditional minority-only oversampling paradigm toward more context-aware, discriminative approaches. Our VAE-GAN+CPAC pipeline offers a principled step in this direction.

\section{Conclusions and Future Works}
\label{sec:conclusions}
This work has demonstrated the limitations of minority-class-only oversampling for fraud detection and highlighted the benefits of classifier-guided, class-aware latent shaping using the CPAC within a VAE-GAN framework. Our results show that relying solely on synthetic augmentation of the minority class can lead to overconfident, poorly generalized models, whereas supervised feedback and prototype-driven clustering provide more meaningful separation and robust detection performance, as reflected in improved F1-score, recall, and AUC.
The CPAC approach offers unique interpretability and flexibility, allowing both effective visualization and strong performance under extreme class imbalance. These findings advocate for a shift away from traditional oversampling techniques towards architectures that explicitly leverage class structure, supervised objectives, and interpretability.
Looking ahead, there are several promising directions for advancing this line of research. Deeper or more complex neural classifiers, either as standalone detectors or as encoder heads, could potentially capture more nuanced fraud patterns. The integration of denoising strategies, such as Denoising Autoencoders, may further enhance confidence and reduce latent overlap. Broader validation across other imbalanced noise detection tasks and domains, as well as the development of richer, more transparent explanation methods, represent important next steps. Finally, exploring alternative strategies for latent space shaping, such as contrastive or manifold-regularized objectives, may yield further gains, especially in low-label or highly imbalanced settings.
Overall, our approach not only advances performance metrics but also addresses the critical needs of interpretability, reliability, and resilience. While focused on credit card fraud, the proposed methods and insights generalize to a wide range of anomaly detection challenges, paving the way for more robust and trustworthy machine learning systems.

\section*{Acknowledgments}
This study has been partially supported by SERICS (PE00000014) under the MUR National Recovery and Resilience Plan funded by the European Union - NextGenerationEU.

\bibliographystyle{unsrt}
\bibliography{references}

\end{document}